%% file: main.tex
\documentclass[sigconf,screen]{acmart}
\usepackage{amsmath}

\usepackage{amssymb}
\usepackage{graphicx}
\usepackage{booktabs}
\usepackage{subcaption}
\usepackage{caption}
\usepackage{multirow}
\usepackage{url}
\usepackage{hyperref}
\usepackage{algorithm}
\usepackage{algorithmic}
\usepackage{pifont}
\usepackage{bbding}
\usepackage{siunitx}     
\usepackage[table]{xcolor}
\definecolor{first}{RGB}{191, 225, 201} 
\definecolor{second}{RGB}{227, 237, 185} 
\definecolor{third}{RGB}{254, 250, 194} 

\renewcommand\footnotetextcopyrightpermission[1]{}

\keywords{Multimodal Image Fusion; Diffusion Transformers; Medical Image Processing} 

\title{MIND: Multimodal Intent-Driven Network via Diffusion Transformers for Medical Image Fusion}

\author{
    Yunzhan Fu\textsuperscript{1} \quad
    Xiangyu Shen\textsuperscript{4} \quad
    Yifei Sun\textsuperscript{2,3}
    \quad
    Yuhan Chen\textsuperscript{5}
    \\
    Jian Wu\textsuperscript{2}$^\dagger$  \quad
    Hongxia Xu\textsuperscript{6}$^\dagger$
}
\affiliation{
    \textsuperscript{1}Transvascular Implantation Devices Research Institute, Zhejiang University, Hangzhou, China \\
    \textsuperscript{2}Zhejiang University, Hangzhou, China \\
    \textsuperscript{3}Liangzhu Laboratory, Hangzhou, China \\
    \textsuperscript{4}Hangzhou Institute of Technology, Xidian University, Hangzhou, China\\
    \textsuperscript{5}Hangzhou Dianzi University, Hangzhou, China \\
    \textsuperscript{6}Zhejiang Key Laboratory of Medical Imaging Artificial Intelligence, Hangzhou
    \country{China}
}

\thanks{$^\dagger$Corresponding authors.}
\begin{abstract}
Medical image fusion aims to integrate complementary information from diverse imaging modalities to support clinical diagnosis. Existing methods typically apply uniform fusion rules globally, lacking a deep understanding of diagnostic intents and pathological structures. To address these limitations, we propose MIND, a Multimodal Intent-Driven Network via Diffusion Transformers (DiTs) for medical image fusion. Specifically, we utilize BioMedGPT to generate intent-driven fusion texts from source images, guiding the fusion process with pathology-aware diagnostic intents. To combat the loss of 2D spatial continuity caused by 1D sequence flattening in DiTs, we design a Multi-scale Latent Adapter. This module explicitly extracts source image  features before serialization, injecting them into the network via strict dimensional alignment to effectively supplement image features. To resolve the semantic shift caused by decoupling image outputs from diagnostic intents, we design a medical semantic consistency loss. This loss ensures deep semantic locking between fused images and fusion texts while maintaining the stability of the underlying physical manifold reconstruction. Comprehensive experiments on the Harvard, BraTS, and GFP datasets reveal that MIND delivers superior fusion quality, significantly improves downstream brain tumor segmentation accuracy, and enables flexible interactive fusion, holding significant promise for intent-driven intelligent clinical decision support systems.
\end{abstract}

\begin{document}
\begin{teaserfigure}
\centering
\includegraphics[width=\textwidth]{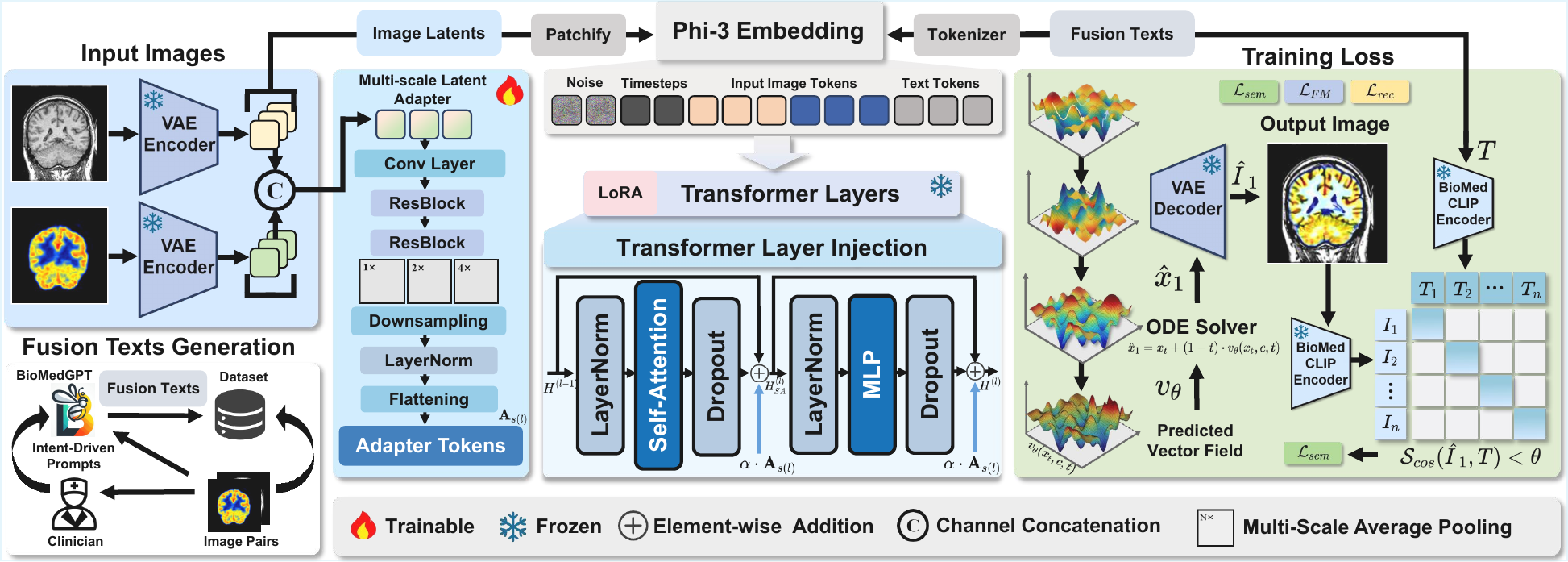}
\captionof{figure}{\textbf{The overall architecture of MIND.} Guided by pathology-aware fusion texts, MIND performs multimodal medical image fusion within a Diffusion Transformers framework.}
\label{fig_overview}
\end{teaserfigure}

\maketitle
\renewcommand{\shortauthors}{Yunzhan Fu et al.}

\begin{figure}[thb]
\centering
\includegraphics[width=1.0\linewidth]{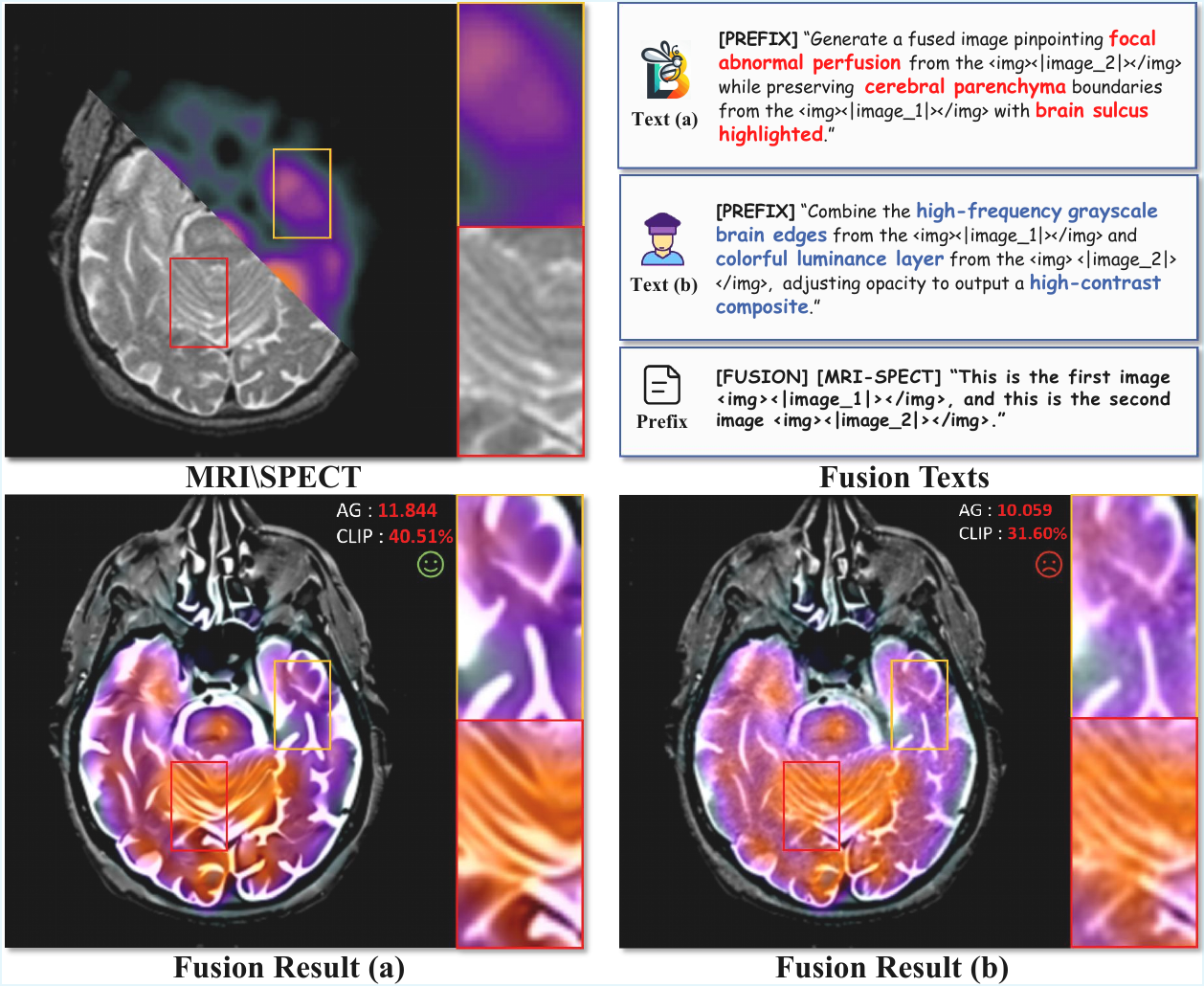}
\vspace{-10pt}
\caption{Fusion Result (a): Intent-driven pathology-aware fusion texts with our MIND framework. Fusion Result (b): Process-driven unified fusion texts with DiTFuse \cite{Li2025DiTFuse}. Our method produces a fused image containing richer gradient information and semantic consistency.
}
\vspace{-10pt}
\label{fig_teaser}
\end{figure}

\section{Introduction}
Medical imaging plays a pivotal role in modern healthcare, disease diagnosis, and treatment planning \cite{HERMESSI2021review}. Due to their distinct physical principles, diverse imaging modalities capture complementary physiological and anatomical information. With the continued advancement of medical imaging technologies, modalities such as magnetic resonance imaging (MRI), computed tomography (CT), and positron emission tomography (PET) provide rich information for clinical diagnosis. However, due to differences in their underlying physics and imaging targets, single-modality images cannot fully represent critical physiological signals and pathological findings \cite{tan2025multimodal}. For instance, MRI offers excellent soft-tissue contrast and clearly delineates the anatomical structures of parenchymal organs \cite{Yang2025TMI}, yet it struggles to detect cortical bone or calcified foci. In contrast, CT excels at delineating osseous structures and calcified lesions with its superior spatial resolution \cite{poornima2025medical}, though its soft-tissue discrimination remains limited. These complementary characteristics underscore the necessity of medical image fusion, which integrates multimodal images to generate a composite image that encompasses both structural and functional information, thereby significantly enhancing diagnostic accuracy \cite{JAMES20144survey}.

The rapid development of deep learning has driven significant progress in image fusion. CNN-based methods \cite{ZHANG202099} leverage their powerful feature extraction capabilities to capture fine-grained textures at multiple scales, yet their limited local receptive fields constrain performance. Subsequent research introduced GAN-based approaches \cite{MA2019FusionGan}, which explicitly enhance fused image quality through the adversarial game between a generator and a discriminator. However, such generative methods often suffer from training instability and a lack of interpretability \cite{ZHOU2023gan}. Moreover, these methods inevitably falter in degraded scenarios. In essence, prior approaches prioritize multimodal information integration without considering effective degradation-aware information recovery \cite{Zhao_2023_CVPR}. Recently, diffusion-based paradigms have demonstrated strong potential in visual restoration tasks within this domain. Existing diffusion-based fusion methods \cite{Zhao_2023_ICCV} typically generate complete images by progressively denoising from noise in an unconstrained latent space \cite{Yue2023DifFusion}. Lacking rigid physical constraints, they tend to fabricate textures in smooth tissues or distort lesion boundaries \cite{He2025TMM}, thereby undermining the pixel-level spatial correspondence essential for downstream clinical diagnosis and high-precision lesion segmentation. Recent studies attempt to explicitly capture complex cross-modal dependencies through multi-scale feature interaction by designing various dedicated image fusion methods \cite{Li2026JBHI}. These customized models are often constrained by specific modality combinations and fixed fusion rules, making it difficult to achieve robust generalization in diverse scenarios. To break the representation bottlenecks of traditional architectures, Diffusion Transformers (DiTs) \cite{sun2026wdt} gradually become the new standard for unified generative paradigms due to their excellent scalability and global context modeling capabilities \cite{Huang_2025_ICCV}. Although the DiT-based paradigms \cite{Cao_2025_ICCV} demonstrate immense potential in vision tasks, their core mechanism of flattening two-dimensional latent variables into 1D sequences for self-attention modeling fundamentally disrupts the local physical continuity required by medical imaging \cite{Xia2025TETCI}.

In real diagnostic scenarios, applying a uniform fusion strategy may weaken the focus on specific pathological targets. The research community has begun to explore semantic-driven image fusion methods, aiming to guide networks to dynamically focus on specific anatomical structures or abnormal metabolic regions by introducing text prompts or prior knowledge \cite{Liu2025JAS}. Recent methods \cite{CHENG2025102790, Yi_2024_CVPR, NEURIPS2024_zhang} explore the use of user instructions to flexibly control the fusion output for different inputs. These methods achieve remarkable results due to the strong representation and generation capabilities of the underlying models. Compared to single-modality image editing tasks, achieving fine-grained and flexible control in image fusion based on semantic priors is highly challenging. DiTFuse \cite{Li2025DiTFuse} proposes a promising unified framework that leverages the strong image reconstruction and semantic understanding capabilities of Omnigen \cite{xiao2025omnigen} to unify various image tasks into a flow matching framework. However, weighting down less relevant content by computing text queries against image keys is an implicit soft re-weighting at the feature level. This lacks explicit constraints on the pixel domain of the final generated image, which easily leads to severe semantic drift and clinical feature loss during highly complex medical feature interactions.

To address these challenges, we propose MIND, a Multimodal Intent-Driven Network via Dits for medical image fusion. As shown in Fig. \ref{fig_teaser}, our approach achieves higher fusion quality and precise clinical semantic control. The main contributions of this paper are summarized as follows:\begin{itemize}
\item We propose a novel medical image fusion network by integrating text-image interaction and visual feature injection within a unified framework. We construct intent-driven fusion texts containing pathological structural elements and descriptions of desired fusion outcomes to guide the training process, thereby significantly enhancing its semantic understanding capability.
\item We design a Multi-scale Latent Adapter (MLA) that explicitly extracts anatomical structure features from input images in latent space and injects them into the DiTs network through dimension alignment, effectively bridging the representational gap between sequence modeling and anatomical topology.
\item We revisit the semantic alignment rules under the DiTs architecture and the flow matching objective, and propose a multimodal medical semantic consistency loss based on a timestep truncation mechanism. This loss ensures deep semantic alignment between the fused image and fusion texts while maintaining absolute stability in the reconstruction of underlying physical structures.
\item Comprehensive experiments demonstrate that our MIND achieves superior fusion quality across various datasets, providing solid visual foundation for downstream task. Moreover, its interactive ability holds promise for future intent-driven intelligent clinical decision support systems.
\end{itemize}

\section{Related Work}
\subsection{Deep Learning-based Medical Image Fusion}
Deep learning-based medical image fusion methods can be categorized into CNN-based, Transformer-based, GAN-based, and Diffusion-based methods. Liu et al. pioneered CNNs in this field, generating fused images that better align with human visual perception \cite{Liu2017CNN}. Zhang et al. proposed IFCNN \cite{ZHANG202099}, a general convolutional framework adapting to multimodal medical, infrared-visible, and multi-focus fusion scenarios. However, limited local receptive fields inherently restrict these architectures from effectively capturing global context and long-range dependencies. To address this limitation, MACTFusion \cite{Xie2025MACTFusion} utilizes cross-attention and transformer modules to model complementary inter-modality relationships. While this yields clearer textures and improved visual quality, substantial computational complexity limits its performance across diverse scenarios. To enhance visual realism, MedFusionGAN \cite{safari2023medfusiongan} refines modality perception and feature selection through generator-discriminator adversarial mechanisms. Nevertheless, typical GAN training instability often causes semantic inconsistency in cross-modal interactions. Diffusion models provide a new generative paradigm for image fusion. DDFM \cite{Zhao_2023_ICCV} first applied them to this task, but its pixel-level similarity objective inherently favors average fusion. Existing uniform fusion rules struggle to balance rigid physical fidelity with complex semantic alignment, thus necessitating a model with fine-grained intent understanding and spatial structural constraints to achieve intent-driven and faithful fusion.

\begin{figure}[thb]
\centering
\includegraphics[width=1.0\linewidth]{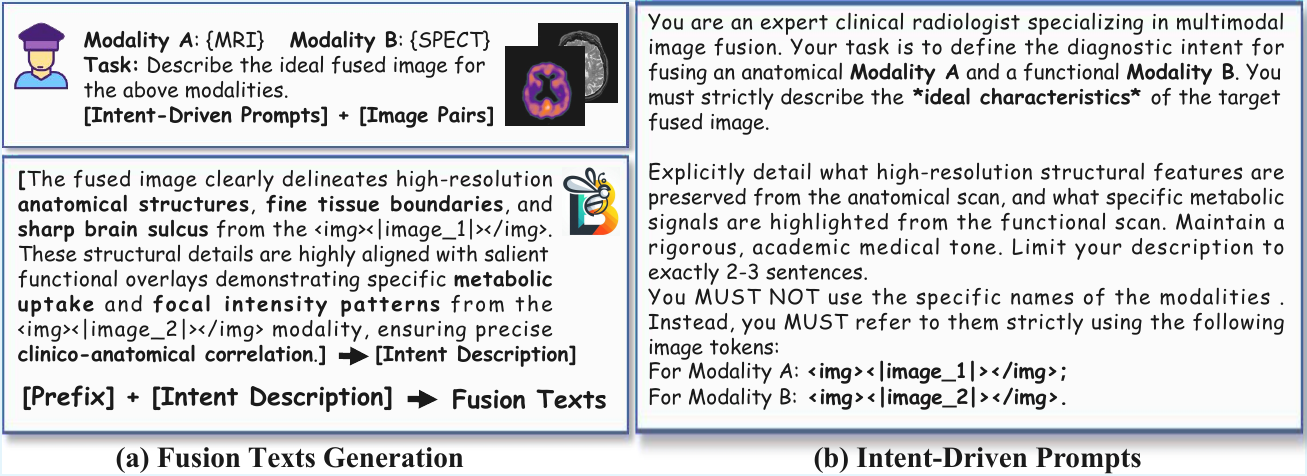}
\vspace{-20pt}
\caption{Detailed prompts for fusion texts generation.}
\vspace{-10pt}
\label{fig:prompts}
\end{figure}

\subsection{Text-Driven Medical Image Fusion}
Text-driven image fusion continuously evolves from general guidance to domain-specific frameworks. As pioneering efforts, Textfusion \cite{CHENG2025102790} achieves feature-level modality integration via affine fusion units, while Text-IF \cite{Yi_2024_CVPR} realizes text-adaptive fusion by combining CLIP embeddings with adaptive instance normalization. Furthermore, FILM \cite{ZHAO2024icml} employs cross-modal attention mechanisms guided by large vision-language models. To enable deeper cross-modal interaction, recent studies explore multi-text collaborative guidance. Specifically, MTG-Fusion \cite{wang2025multi} leverages multi-dimensional descriptions to precisely screen shared features and adaptively fuse modality-specific ones. Concurrently, DiTFuse \cite{Li2025DiTFuse} adopts a DiTs backbone to process text instructions and images in parallel, thereby achieving end-to-end control through iterative denoising. Despite these advancements, existing methods predominantly utilize text to describe source features or the fusion process, termed Process-driven, which limits their ability to foresee specific diagnostic goals.

\section{Methodology}
\label{sec:method}
\subsection{Overview}
\label{subsec:overview}
As illustrated in Fig. \ref{fig_overview}, MIND adopts a solid foundation that utilizes the powerful reconstruction ability from OmniGen \cite{xiao2025omnigen}. Given a pair of co-registered anatomical $I_A \in \mathbb{R}^{1 \times H \times W}$ and functional $I_B \in \mathbb{R}^{3 \times H \times W}$ medical images, we first leverage BioMedGPT \cite{Zhang2024BiomedGPT} to formulate explicit diagnostic guidance. Unlike \cite{Xiang25JBHI} that passively describes source inputs, our prompt engineering is strictly Intent-Driven. As shown in Fig. \ref{fig:prompts} (b), we instruct the model to foresee the ideal characteristics of the target fused image. This generated intent description is concatenated with a standard prefix in Fig. \ref{fig_teaser} to form the fusion texts $T$. Our general objective is to learn a conditional probability flow model to ensure that $I_{fused}$ strictly preserves anatomical topologies while accurately aligning with high-level clinical intentions. The generation process can be formulated as:
\begin{equation}
    I_{fused} = \mathcal{F}_\theta(x_0 \mid I_A, I_B, T), \quad x_0 \sim \mathcal{N}(0, \mathbf{I})
\end{equation}
where $\mathcal{F}_\theta$ denotes the continuous flow transformation that maps an initial Gaussian noise distribution $x_0$ to the target data distribution, and $\mathbf{I}$ represents the identity matrix matching the latent dimensions.

\subsection{Model Architecture and Data Input}
\label{subsec:model_design}
Unlike traditional convolutional networks operating in the computationally expensive pixel space, our core fusion network comprises 32 stacked Phi-3 Transformer blocks. Specifically, we utilize a pre-trained frozen Variational Autoencoder (VAE) $\mathcal{E}_{vae}$ from SDXL \cite{podell2023SDXLVAE} to encode the heterogeneous source images $I \in \{I_A, I_B\}$ into compact latent representations:
\begin{equation}
    x_A = \mathcal{E}_{vae}(I_A) \in \mathbb{R}^{C \times \frac{H}{8} \times \frac{W}{8}}
\end{equation}
\begin{equation}
    x_B = \mathcal{E}_{vae}(I_B) \in \mathbb{R}^{C \times \frac{H}{8} \times \frac{W}{8}}
\end{equation}
where $C$ denotes the latent channel dimension. For fusion texts $T$, we leverage the Phi-3 tokenizer \cite{abdin2024phi3} and its embedding layer to transform $T$ into a contextualized semantic sequence $\mathbf{E}_T$. 

To construct the joint input sequence, the image latents $x_A$ and $x_B$ are patchified, flattened, and linearly projected into visual tokens $\mathbf{X}_A$ and $\mathbf{X}_B$. As shown in Fig. \ref{fig:prompts}, we follow the operation in \cite{Li2025DiTFuse} to encapsulate these visual tokens with special \texttt{<img>} and \texttt{</img>} tokens and subsequently concatenate them with the text embedding $\mathbf{E}_T$ to generate text-image condition context embedding $\mathbf{E}_c$. At the input layer of the Phi3 Transformer \cite{abdin2024phi3} backbone, the text-image condition context embedding $\mathbf{E}_c$, the timestep embedding $\mathbf{E}_t$, and the noisy target image tokens $\mathbf{X}_t$ are concatenated into an initial joint input sequence $H^{(0)} = [\mathbf{E}_c, \mathbf{E}_t, \mathbf{X}_t]$.

\subsection{Multi-scale Latent Adapter}
\label{subsec:adapter}
Inspired by T2I-Adapter \cite{Mou2024adapter}, which seamlessly integrates external spatial guidance into frozen pre-trained diffusion models, we designed MLA. Image latents $x_A$ and $x_B$ are processed with channel concatenation to form the input adapter latent $x_{in}$. The MLA network $\Phi_{adapt}$ utilizes multi-scale residual blocks in the 2D domain to explicitly extract continuous anatomical edges before serialization. These locally encoded 2D spatial priors are subsequently downsampled via different pooling strides to construct a hierarchical feature pyramid. After aligning with the target sequence dimension, the flattened anatomical prior sequence for the $s$-th scale is denoted as:
\begin{equation}
    F_{adapt}^{(s)} = \text{Flatten}\Big(\text{LN}\big(\text{Conv}_{P \times P}\big(\Phi_{adapt}^{(s)}(x_{in})\big)\big)\Big) \in \mathbb{R}^{L_s \times D}
\end{equation}
where $\text{LN}(\cdot)$ denotes Layer Normalization. $L_s$ is the sequence length of the $s$-th scale, and $D=3072$ is the hidden dimension of the Transformer.

To robustly inject the multi-scale features into the $N=32$ layers of the Transformer without disrupting the pre-trained knowledge, we pad $F_{adapt}^{(s)}$ to the exact sequence length of $\mathbf{E}_c$, obtaining $\tilde{F}_{adapt}^{(s)}$. Then, we construct aligned adapter sequences $\mathbf{A}_s = [\tilde{F}_{adapt}^{(s)}, \mathbf{0}_t, \mathbf{0}_x]$ that match the length and positional indices of $H^{(0)}$. Following Transformer dynamics \cite{raghu2021vision}, where shallow layers capture high-frequency details and deep layers aggregate global semantics, we adopt a linear allocation strategy that maps the scale index $s \in \{0, \dots, S-1\}$ to the transformer layer index $l \in \{0, \dots, N-1\}$:
\begin{equation}
    s(l) = \min\left(\left\lfloor \frac{l \cdot S}{N} \right\rfloor, S - 1\right)
\end{equation}
For $S=3$ and $N=32$, the specific allocation is: $l=0\sim10$ for scale-0, $l=11\sim21$ for scale-1, and $l=22\sim31$ for scale-2. We conducted detailed analysis of this machanism in the Supplementary Material. 

During the forward pass within the $l$-th hidden layer, the corresponding anatomical prior $\mathbf{A}_{s(l)}$ is seamlessly injected as a residual term:
\begin{equation}
    H^{(l)}_{SA} = \text{SelfAttention}\big(\text{LN}(H^{(l-1)})\big) + H^{(l-1)} + \alpha \cdot \mathbf{A}_{s(l)}
\end{equation}
\begin{equation}
    H^{(l)} = \text{MLP}\big(\text{LN}(H^{(l)}_{SA})\big) + H^{(l)}_{SA} + \alpha \cdot \mathbf{A}_{s(l)}
\end{equation}
where $\alpha$ is a hyperparameter for scaling.

\subsection{Loss Function and Training Strategy}
\label{loss_function}
To enable efficient conditional generation, we adopt Continuous Flow Matching (CFM) \cite{lipman2023flowmatching} as the foundational generative dynamics. The generation process is defined as an Ordinary Differential Equation (ODE) trajectory from a Gaussian noise prior $x_0 \sim \mathcal{N}(0, \mathbf{I})$ to the target latent distribution $x_1 \sim q(x_1)$. We construct an optimal transport path where the intermediate state $x_t$ at any continuous timestep $t \in [0, 1]$ is defined as:
\begin{equation}
    x_t = t x_1 + (1 - t) x_0
\end{equation}
The corresponding target vector field is a constant $u_t = x_1 - x_0$. The unified generative backbone $\theta$ receives the fusion texts embedding $\mathbf{E}_T$, the timestep embedding $\mathbf{E}_t$, and the noisy image tokens $\mathbf{X}_t$. Its optimization objective is to minimize the mean squared error between the predicted vector field $v_\theta$ and the target field $u_t$:
\begin{equation}
    \mathcal{L}_{FM} = \mathbb{E}_{t, x_0, x_1, T} \left[ \| v_\theta(x_t, T, t) - (x_1 - x_0) \|_2^2 \right]
\end{equation}

While $\mathcal{L}_{FM}$ constrains the global manifold reconstruction, it cannot guarantee fine-grained alignment between the generated medical details and the fusion texts $T$. To address this optimization dilemma, we propose a Multimodal Medical Semantic Consistency Loss based on a timestep-truncation mechanism to avoid gradient conflicts. At any given timestep $t$, we estimate the current clean target latent $\hat{x}_1$ using first-order Euler integration:
\begin{equation}
    \hat{x}_1 = x_t + (1 - t) \cdot v_\theta(x_t, T, t)
\end{equation}
Subsequently, $\hat{x}_1$ is mapped back to the pixel domain via the VAE decoder $\mathcal{D}_{vae}$, yielding the estimated image $\hat{I}_1 = \mathcal{D}_{vae}(\hat{x}_1)$. We introduce a Timestep-Truncation Mechanism $\mathbb{I}(t > \tau)$ and a threshold $\theta$, which strictly restricts the activation of semantic penalties. 
We first define the cosine similarity between the generated image and the textual instruction as:
\begin{equation}
    \mathcal{S}_{cos}(\hat{I}_1, T) = \frac{\mathcal{E}_{CLIP-I}(\hat{I}_1) \cdot \mathcal{E}_{CLIP-T}(T)}{\|\mathcal{E}_{CLIP-I}(\hat{I}_1)\| \|\mathcal{E}_{CLIP-T}(T)\|}
\end{equation}
where $\mathcal{E}_{CLIP-I}$ and $\mathcal{E}_{CLIP-T}$ denote the frozen image and text encoders of BioMedCLIP \cite{zhang2025biomedclip}. Based on this, the overall semantic consistency loss is calculated as:
\begin{equation}
    \ell_{sem}(\hat{I}_1, T) = 
    \begin{cases} 
        0, & if \quad \mathcal{S}_{cos}(\hat{I}_1, T) \geq \theta \\ 
        1 - \mathcal{S}_{cos}(\hat{I}_1, T), & otherwise 
    \end{cases}
\end{equation}
\begin{equation}\label{eq:l_sem}
    \mathcal{L}_{sem} = \mathbb{E}_{t \sim \mathcal{U}(0,1)} \Big[ \mathbb{I}(t > \tau) \cdot t \cdot \ell_{sem}(\hat{I}_1, T) \Big]
\end{equation}

To strictly guarantee the clinical reference value of the fused image at the high-frequency level, we superimpose physical reconstruction constraints. Concretely, we apply the $L_1$ norm in the latent space and the Structural Similarity Index Measure (SSIM) \cite{zhao2016loss} in the decoded pixel space:
\begin{equation}\label{eq:l_rec}
    \mathcal{L}_{rec} = \mathbb{E}_{t}\Big[ t \cdot \big( \lambda_{L1} \|\hat{x}_1 - x_A\|_1 + \lambda_{SSIM} (1-\text{SSIM}(\hat{I}_1, I_{A})) \big)\Big]
\end{equation}
As formulated in Eqs. (\ref{eq:l_sem}) and (\ref{eq:l_rec}), both $\mathcal{L}_{sem}$ and $\mathcal{L}_{rec}$ are explicitly scaled by timestep $t$. This time-aware dynamic weighting mechanism suppresses unstable gradients at early, high-noise stages to preserve the underlying physical skeleton \cite{Hang_2023_ICCV}. Conversely, it amplifies penalties during late stages to enforce strict alignment of high-frequency textures and semantics. The overall multi-objective optimization function is formulated as:
\begin{equation}
    \mathcal{L}_{total} = \mathcal{L}_{FM} + \mathcal{L}_{rec} + \lambda_{sem} \mathcal{L}_{sem}
\end{equation} 
where $\lambda_{L1}$, $\lambda_{SSIM}$, and $\lambda_{sem}$ are wights balancing the respective loss terms. The network is fine-tuned by introducing Low-Rank Adaptation (LoRA) \cite{hu2022lora} into the linear layers of the transformer blocks, alongside the full-parameter updating of MLA.

\begin{figure*}[thb]
\centering
\includegraphics[width=1.0\linewidth]{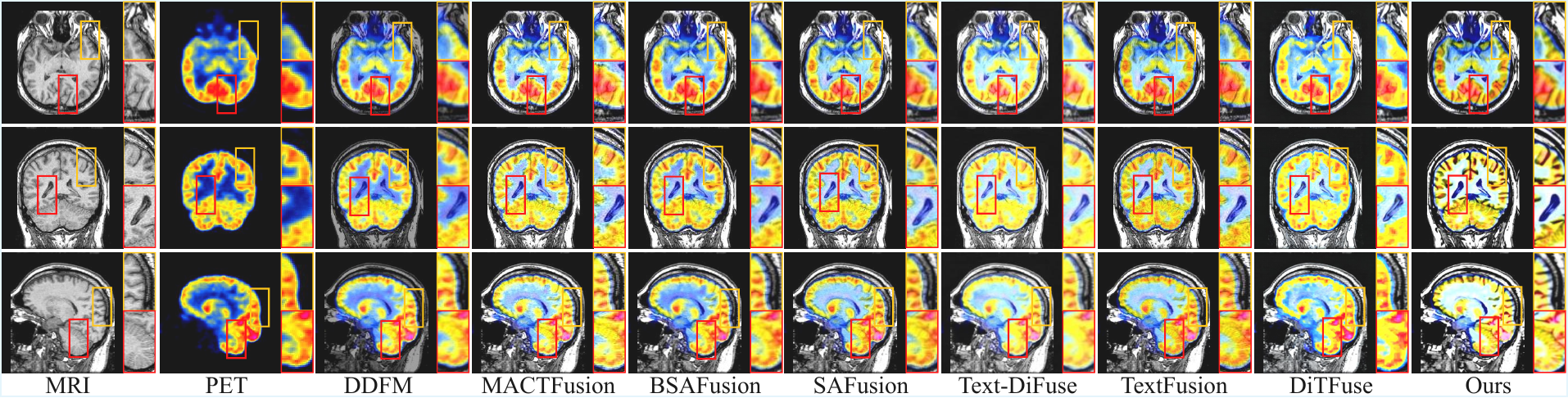}
\vspace{-20pt}
\caption{PET-MRI fusion results on Harvard. Red and orange boxes highlight fine-grained local details.}
\vspace{-10pt}
\label{fig_PET-MRI}
\end{figure*}

\begin{figure*}[thb]
\centering
\includegraphics[width=1.0\linewidth]{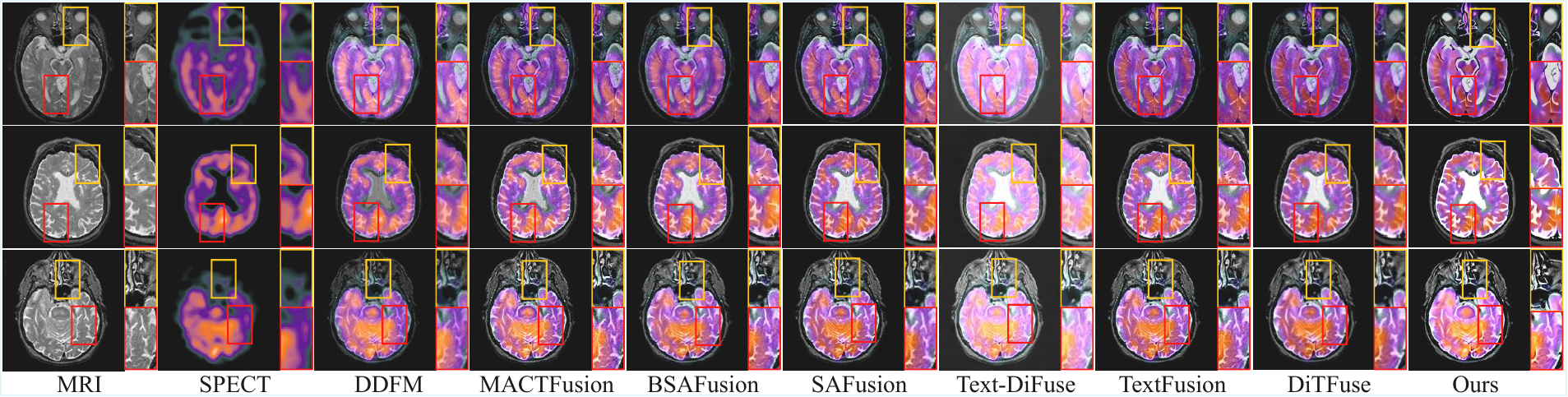}
\vspace{-20pt}
\caption{SPECT-MRI fusion results on Harvard. Red and orange boxes highlight fine-grained local details.}
\vspace{-10pt}
\label{fig_SPECT-MRI}
\end{figure*}

\begin{figure*}[thb]
\centering
\includegraphics[width=1.0\linewidth]{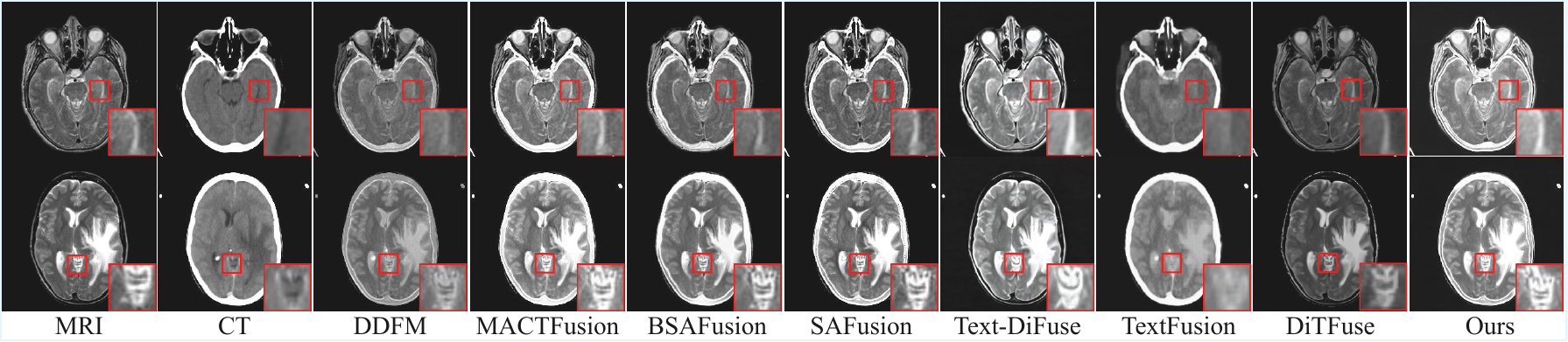}
\vspace{-20pt}
\caption{CT-MRI fusion results on Harvard. Red boxes highlight fine-grained local details.}
\vspace{-10pt}
\label{fig_CT-MRI}
\end{figure*}

\begin{table*}[!t]
\centering
\caption{Quantitative comparison of MIND with eight state-of-the-art methods on Harvard. Best results are highlighted as \colorbox{first}{\textbf{first}}, \colorbox{second}{second} and \colorbox{third}{third}.}
\vspace{-10pt}
\label{tab:harvard_quant}
\footnotesize
\setlength{\tabcolsep}{3pt}  
\resizebox{\linewidth}{!}{
\begin{tabular}{lllcccccccc}
\toprule
\textbf{Dataset} & \textbf{Method} & \textbf{Source} & \textbf{EN} & \textbf{AG} & \textbf{MI} & \textbf{VIF} & \textbf{SF} & \textbf{SD} & \textbf{Qabf} & \textbf{MS\_SSIM} \\
\midrule

\multirow{9}{*}{\textbf{CT-MRI}}
    & DDFM \cite{Zhao_2023_ICCV}           & ICCV$_{23}$
    & 4.195±0.196
    & 5.461±0.855
    & 2.675±0.292
    & 0.402±0.039
    & 22.165±3.019
    & 62.305±4.101
    & 0.442±0.041
    & \cellcolor{third}{0.543±0.041} \\

    & Text-DiFuse \cite{NEURIPS2024_zhang} & NeurIPS$_{24}$
    & \cellcolor{second}{5.577±0.255}
    & 8.433±1.244
    & \cellcolor{second}{3.247±0.292}
    & 0.484±0.041
    & 34.606±3.839
    & \cellcolor{third}{87.012±3.490}
    & 0.570±0.024
    & 0.401±0.075 \\

    & FILM \cite{ZHAO2024icml}              & ICML$_{24}$
    & 3.907±0.241
    & \cellcolor{second}{9.593±1.504}
    & \cellcolor{third}{2.986±0.202}
    & 0.403±0.036
    & \cellcolor{first}{\textbf{42.957±6.171}}
    & 83.367±4.030
    & \cellcolor{second}{0.644±0.032}
    & 0.368±0.013 \\

    & MACTFusion \cite{Xie2025MACTFusion}  & JBHI$_{25}$
    & 4.543±0.241
    & 8.066±1.316
    & 2.854±0.155
    & \cellcolor{third}{0.501±0.048}
    & 34.442±4.526
    & 86.456±5.119
    & \cellcolor{third}{0.612±0.041}
    & \cellcolor{second}{0.550±0.048} \\

    & BSAFusion \cite{Li_Su_Cai_Zhang_2025}& AAAI$_{25}$
    & 4.638±0.255
    & 7.276±1.220
    & 2.829±0.263
    & 0.392±0.067
    & 27.964±3.453
    & 80.277±5.046
    & 0.427±0.056
    & 0.469±0.086 \\

    & TextFusion \cite{CHENG2025102790}    & INFFUS$_{25}$
    & \cellcolor{third}{4.775±0.248}
    & 5.020±0.960
    & 2.944±0.197
    & 0.436±0.038
    & 22.424±2.936
    & 81.132±6.260
    & 0.380±0.032
    & 0.500±0.037 \\

    & DiTFuse \cite{Li2025DiTFuse}         & TPAMI$_{25}$
    & 3.884±0.284
    & 8.979±1.353
    & 2.319±0.309
    & 0.412±0.864
    & 32.050±4.044
    & \cellcolor{second}{87.438±13.169}
    & 0.471±0.061
    & 0.348±0.102 \\

    & SAFusion \cite{Li2026JBHI}            & JBHI$_{26}$
    & 4.533±0.236
    & \cellcolor{third}{9.119±1.443}
    & 2.832±0.230
    & \cellcolor{first}{\textbf{0.581±0.0095}}
    & \cellcolor{third}{38.610±5.084}
    & 82.766±4.991
    & 0.348±0.043
    & \cellcolor{first}{\textbf{0.573±0.0122}} \\

    & \textbf{MIND}                        & \textbf{Ours}
    & \cellcolor{first}{\textbf{5.802±0.149}}
    & \cellcolor{first}{\textbf{10.111±2.243}}
    & \cellcolor{first}{\textbf{3.274±0.201}}
    & \cellcolor{second}{0.569±0.06}
    & \cellcolor{second}{41.250±7.704}
    & \cellcolor{first}{\textbf{92.291±6.454}}
    & \cellcolor{first}{\textbf{0.672±0.0176}}
    & 0.488±0.092 \\
\midrule

\multirow{9}{*}{\textbf{PET-MRI}}
    & DDFM \cite{Zhao_2023_ICCV}           & ICCV$_{23}$
    & 4.922±0.600
    & 7.411±0.930
    & 2.455±0.416
    & 0.614±0.05
    & 23.376±2.120
    & 78.061±4.970
    & 0.571±0.032
    & 0.621±0.021 \\

    & FILM \cite{ZHAO2024icml}              & ICML$_{24}$
    & 3.437±0.835
    & 8.121±3.050
    & \cellcolor{second}{3.746±0.305}
    & \cellcolor{first}{\textbf{0.793±0.074}}
    & 31.297±8.839
    & 72.937±10.140
    & \cellcolor{first}{\textbf{0.784±0.025}}
    & 0.618±0.085 \\

    & Text-DiFuse \cite{NEURIPS2024_zhang} & NeurIPS$_{24}$
    & \cellcolor{second}{6.224±0.534}
    & 10.889±1.373
    & 3.127±0.202
    & 0.526±0.045
    & 35.672±3.362
    & \cellcolor{second}{97.248±3.095}
    & 0.613±0.039
    & \cellcolor{first}{\textbf{0.698±0.028}} \\

    & MACTFusion \cite{Xie2025MACTFusion}  & JBHI$_{25}$
    & 5.430±0.646
    & 10.306±1.192
    & \cellcolor{third}{3.498±0.274}
    & \cellcolor{second}{0.779±0.068}
    & 35.705±3.668
    & 91.352±3.304
    & \cellcolor{third}{0.756±0.036}
    & \cellcolor{third}{0.637±0.016} \\

    & BSAFusion \cite{Li_Su_Cai_Zhang_2025}& AAAI$_{25}$
    & 5.196±0.659
    & 11.551±1.591
    & \cellcolor{third}{3.498±0.304}
    & 0.663±0.052
    & \cellcolor{second}{37.399±4.049}
    & 90.080±2.526
    & 0.699±0.0364
    & 0.626±0.022 \\

    & TextFusion  \cite{CHENG2025102790}   & INFFUS$_{25}$
    & \cellcolor{third}{5.770±0.641}
    & 9.983±1.345
    & 3.345±0.244
    & 0.633±0.023
    & 32.018±3.335
    & 83.078±3.279
    & 0.696±0.029
    & 0.603±0.016 \\

    & DiTFuse \cite{Li2025DiTFuse}         & TPAMI$_{25}$
    & 4.868±0.669
    & \cellcolor{first}{\textbf{12.423±1.838}}
    & 2.752±0.248
    & 0.557±0.239
    & 33.819±2.929
    & \cellcolor{third}{94.810±8.598}
    & 0.640±0.052
    & 0.557±0.033 \\

    & SAFusion \cite{Li2026JBHI}            & JBHI$_{26}$
    & 5.298±0.605
    & \cellcolor{third}{11.719±1.482}
    & 2.495±0.132
    & 0.661±0.041
    & \cellcolor{third}{35.907±5.084}
    & 87.109±2.727
    & \cellcolor{second}{0.759±0.037}
    & 0.619±0.018 \\

    & \textbf{MIND}                        & \textbf{Ours}
    & \cellcolor{first}{\textbf{6.238±0.526}}
    & \cellcolor{second}{11.893±1.539}
    & \cellcolor{first}{\textbf{4.063±0.359}}
    & \cellcolor{third}{0.749±0.118}
    & \cellcolor{first}{\textbf{42.302±6.473}}
    & \cellcolor{first}{\textbf{97.716±2.907}}
    & 0.750±0.096
    & \cellcolor{second}{0.642±0.028} \\
\midrule

\multirow{9}{*}{\textbf{SPECT-MRI}}
    & DDFM \cite{Zhao_2023_ICCV}           & ICCV$_{23}$
    & 4.392±0.473
    & 5.198±1.014
    & \cellcolor{third}{3.262±0.217}
    & 0.477±0.04
    & 17.399±2.378
    & 64.194±5.449
    & 0.569±0.044
    & 0.589±0.090 \\

    & FILM \cite{ZHAO2024icml}              & ICML$_{24}$
    & 3.776±0.493
    & 5.728±1.227
    & 2.807±0.289
    & \cellcolor{third}{0.754±0.124}
    & 22.687±4.865
    & \cellcolor{third}{66.038±10.784}
    & 0.671±0.036
    & \cellcolor{second}{0.648±0.105} \\

    & TextFusion \cite{CHENG2025102790}    & INFFUS$_{25}$
    & \cellcolor{third}{5.032±0.631}
    & 6.103±1.167
    & \cellcolor{second}{3.444±0.280}
    & 0.672±0.050
    & 20.240±2.626
    & 60.185±5.237
    & 0.677±0.044
    & 0.585±0.074 \\

    & MACTFusion \cite{Xie2025MACTFusion}  & JBHI$_{25}$
    & 4.994±0.707
    & 6.969±1.248
    & 3.129±0.273
    & 0.690±0.045
    & 22.018±2.927
    & 60.287±5.380
    & \cellcolor{third}{0.730±0.035}
    & \cellcolor{third}{0.621±0.050} \\

    & BSAFusion \cite{Li_Su_Cai_Zhang_2025}& AAAI$_{25}$
    & 4.643±0.539
    & 6.927±1.402
    & 3.169±0.255
    & 0.625±0.046
    & \cellcolor{third}{23.623±3.393}
    & 64.067±5.367
    & 0.678±0.050
    & 0.609±0.065 \\

    & Text-DiFuse \cite{NEURIPS2024_zhang} & NeurIPS$_{24}$
    & \cellcolor{second}{5.895±0.390}
    & \cellcolor{first}{\textbf{7.196±1.583}}
    & 3.074±0.161
    & 0.547±0.033
    & \cellcolor{first}{\textbf{26.081±4.273}}
    & \cellcolor{second}{77.622±5.043}
    & 0.584±0.046
    & 0.486±0.030 \\

    & DiTFuse \cite{Li2025DiTFuse}         & TPAMI$_{25}$
    & 4.275±0.470
    & 6.947±1.252
    & 2.616±0.202
    & 0.570±0.049
    & \cellcolor{second}{24.304±2.754}
    & 63.123±7.982
    & 0.553±0.045
    & 0.515±0.083 \\

    & SAFusion \cite{Li2026JBHI}            & JBHI$_{26}$
    & 4.815±0.614
    & \cellcolor{second}{7.145±1.231}
    & 2.986±0.144
    & \cellcolor{second}{0.835±0.082}
    & 22.576±2.825
    & 61.754±4.600
    & \cellcolor{second}{0.756±0.037}
    & 0.614±0.063 \\

    & \textbf{MIND}                        & \textbf{Ours}
    & \cellcolor{first}{\textbf{6.236±0.419}}
    & \cellcolor{third}{6.974±1.333}
    & \cellcolor{first}{\textbf{4.142±0.587}}
    & \cellcolor{first}{\textbf{0.877±0.035}}
    & 23.316±3.029
    & \cellcolor{first}{\textbf{79.905±4.388}}
    & \cellcolor{first}{\textbf{0.792±0.052}}
    & \cellcolor{first}{\textbf{0.659±0.026}} \\
\bottomrule
\end{tabular}
}
\end{table*}

\section{Experiments}
\subsection{Dataset and Evaluation Metrics}
\noindent\textbf{Data Preparation.}
To comprehensively evaluate the effectiveness of our framework, we curated a diverse and highly representative experimental dataset from three publicly available sources. All image pairs are rigidly co-registered prior to the experiments. 

\noindent\textbf{Harvard Whole Brain Atlas (Harvard)} constitutes a comprehensive medical imaging resource. Specifically, we extracted 160 CT-MRI pairs, 245 PET-MRI pairs, and 333 SPECT-MRI pairs for training set. We selected 24 pairs from each category for testing.

\noindent\textbf{The Brain Tumor Segmentation Dataset (BraTS)} \cite{brats_dataset} is an authoritative public resource for multimodal magnetic resonance imaging analysis. It contains brain imaging data across scanning sequences including T1, T1ce, T2, and FLAIR. We used BraTS 2017, which comprises 484 clinical cases. Each case contains 155 slices with a $240 \times 240$ pixel resolution and includes corresponding label masks delineating peritumoral Edema (ED), Enhancing Tumor (ET), and Non-Enhancing Tumor regions (NET). T1ce emphasizes the active tumor core via contrast enhancement, whereas FLAIR suppresses cerebrospinal fluid to better delineate edema and non-enhancing tumor regions. We assign 387 T1-FLAIR pairs to the training set and the remaining 97 pairs to the test set.

\noindent\textbf{GFP Dataset (GFP)} \cite{koroleva2005high} is a multimodal cellular imaging collection comprising 155 pairs of Green Fluorescent Protein and Phase Contrast (GFP-PC) microscopy images. The fluorescence images deliver detectable signals, while the phase contrast images supply the structural background of the cells. We selected 148 GFP-PC pairs for external validation to assess the generalization capacity.

\noindent\textbf{Data Pre-processing.} The BraTS dataset consists of 3D volumetric scans. To adapt it to the input structure of 2D vision models, we utilized an information-guided dimensionality reduction method. For each 3D sample, we traversed a specific axis and computed the tumor area in the corresponding mask $M_i$ for each slice $S_i$. The optimal 2D slice $S^*$ was selected as $S^* = \arg\max_{S_i} \operatorname{Area}(M_i)$. Through this process, we extracted the core slice with the maximum tumor spatial occupancy and richest semantic information, along with its label, constructing a 2D multimodal dataset. Furthermore, to ensure dimensional consistency, we uniformly resized the above datasets to a resolution of $256 \times 256$ pixels.

\noindent\textbf{Evaluation Metrics.}
To quantitatively evaluate the fusion results, we employ eight widely used metrics: Spatial Frequency (SF) \cite{shapley1985spatial}, Entropy (EN) \cite{bein2006entropy}, Average Gradient (AG) \cite{chaithra2018survey}, Mutual Information(MI) \cite{kraskov2004estimating}, Visual Information Fidelity (VIF) \cite{han2013new}, Standard Deviation (SD) \cite{Electric2000letter}, Quality assessment based on fusion (Qab/f) \cite{MA2019153}, and Multi-Scale Structural Similarity Index Measure (MS-SSIM) \cite{wang2003ssim}. Following the wide-adopted CLIP-Score \cite{hessel-etal-2021-clipscore}, we computed the cosine similarity between the fused images and fusion texts using BioMedCLIP \cite{zhang2025biomedclip} to quantify their semantic alignment. The selected metrics comprehensively quantify fusion performance by providing an in-depth evaluation of medical image fusion methods.

\begin{figure*}[thb]
\centering
\includegraphics[width=1.0\linewidth]{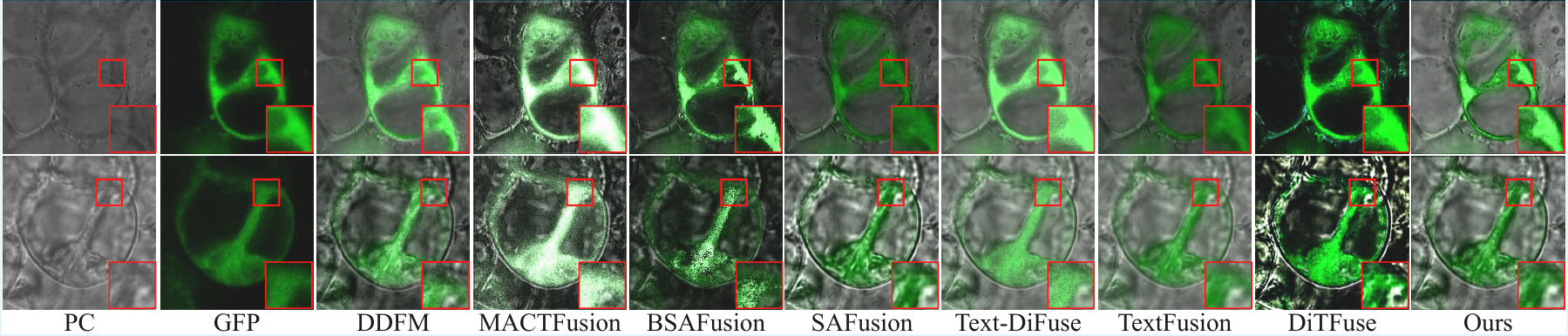}
\vspace{-20pt}
\caption{GFP-PC fusion results on GFP. Prompts are adapted to biological contexts to evaluate cross-domain robustness.}
\vspace{-10pt}
\label{fig_GFP}
\end{figure*}

\subsection{Implementation Details}
We trained our model on an NVIDIA RTX 5090 GPU and the PyTorch 2.12.0 framework. The network optimization was carried out with AdamW, where $\beta_1$ and $\beta_2$ were set to 0.9 and 0.999, respectively. The initial learning rate was set to $1\times10^{-7}$, and the scheduler was cosine with 1,000 warmup steps. We used a batch size of 2 per GPU and 4 data loader workers. The training ran for 20 epochs with gradient accumulation set to 1. We used LoRA with the rank of 64 for efficient parameter tuning and set conditional dropout probability to 0.01 for modality robustness. Furthermore, the hyperparameters $\alpha$ and $\tau$ were cautiously set to 1.0 and 0.50, respectively. For reconstruction loss, we used $\lambda_{L1}=0.15$ and $\lambda_{SSIM}=0.05$. For semantic loss, $\lambda_{sem}$ and $\theta$ were set to 0.1 and 0.85. We provide careful analysis of these parameters in section \ref{sec:parameter} and Supplementary Material. We conducted both qualitative and quantitative analyses to compare 8 different fusion approaches, including DDFM \cite{Zhao_2023_ICCV}, MACTFusion \cite{Xie2025MACTFusion}, BSAFusion \cite{Li_Su_Cai_Zhang_2025}, SAFusion \cite{Li2026JBHI}, Text-DiFuse \cite{NEURIPS2024_zhang}, TextFusion \cite{CHENG2025102790} , DiTFuse \cite{Li2025DiTFuse} and FILM \cite{ZHAO2024icml}. For text-driven methods \cite{NEURIPS2024_zhang, CHENG2025102790, Li2025DiTFuse, ZHAO2024icml}, we use their officially optimized prompts to ensure their peak performance. We tested different prompts to verify robustness in Supplementary Material. For downstream validation, we conduct brain tumor segmentation experiments on the BraTS 2017 \cite{brats_dataset} using the nnUNet backbone \cite{nnUNet}. We set the initial learning rate to 0.01 with a Polynomial Learning Rate Scheduler, and train the models for 1000 epochs with a batch size of 18.

\subsection{Comparative Experiment}

\subsubsection{Qualitative Comparisons.}

Fig. \ref{fig_PET-MRI}, \ref{fig_SPECT-MRI}, and \ref{fig_CT-MRI} present qualitative comparisons of MIND against state-of-the-art methods across three subsets of Harvard. Fig. \ref{fig_GFP} evaluates cross-dataset generalization on GFP. Fine-grained details within the regions of interest are highlighted with red and orange boxes. In the PET-MRI Fusion, MACTFusion \cite{Xie2025MACTFusion} and TextFusion \cite{CHENG2025102790} suffer from severe texture degradation in magnified regions. For the SPECT-MRI Fusion, Text-DiFuse \cite{NEURIPS2024_zhang} and DiTFuse \cite{Li2025DiTFuse} undergo significant contrast attenuation, indicating an inability to reliably map functional intensities. MIND safely bypasses these bottlenecks, ensuring high spatial fidelity and fine texture structure. For fusing task of CT and MRI, DDFM \cite{Zhao_2023_ICCV} introduces noticeable blurring and reduced contrast. While MACTFusion \cite{Xie2025MACTFusion}, BSAFusion \cite{Li_Su_Cai_Zhang_2025}, and SAFusion \cite{Li2026JBHI} effectively retain MRI soft tissues, they tend to suppress prominent CT features. MIND yields a much more favorable balance, efficiently fusing dense anatomical details from both modalities without introducing visual artifacts or semantic degradation. As for the external GFP, DDFM \cite{Zhao_2023_ICCV}, MACTFusion \cite{Xie2025MACTFusion}, BSAFusion \cite{Li_Su_Cai_Zhang_2025}, and DiTFuse \cite{Li2025DiTFuse} preserve color information but suffer from severe structural degradation. Compared with SAFusion \cite{Li2026JBHI} and TextFusion \cite{CHENG2025102790}, MIND achieves a better visual balance, effectively maintaining both functional color maps and fine-grained textures. These results demonstrate MIND’s robust cross-modal alignment and strong generalization across diverse data distributions.
\begin{figure*}[thb]
\centering
\includegraphics[width=1.0\linewidth]{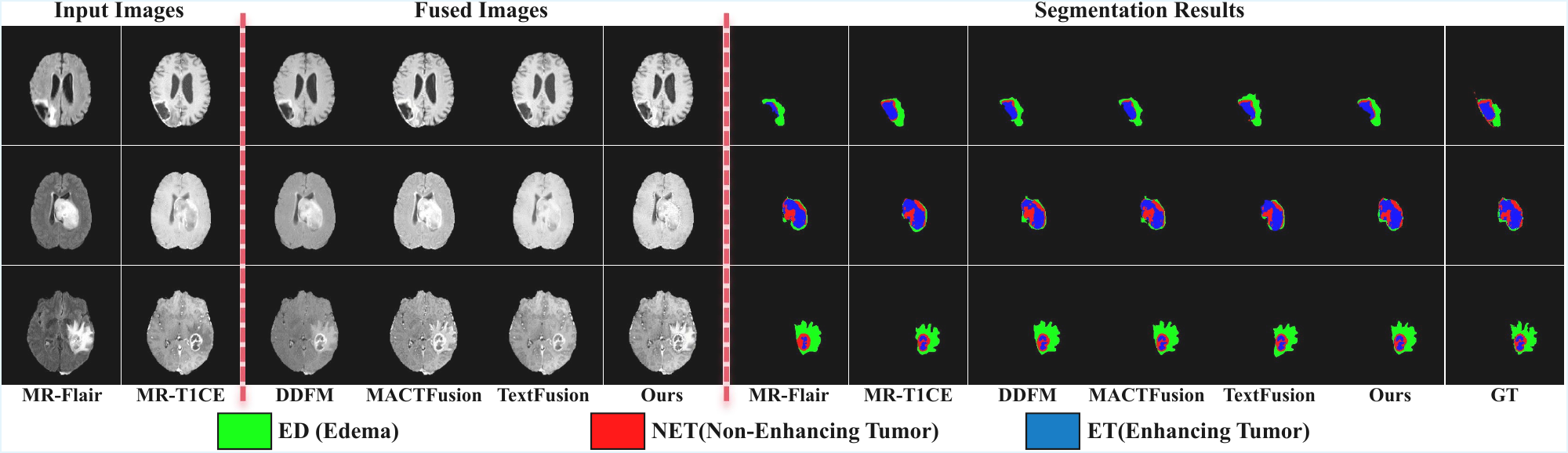}
\vspace{-20pt}
\caption{Brain tumor segmentation results using nnUNet \cite{nnUNet} on BraTS 2017 \cite{brats_dataset}.}
\vspace{-10pt}
\label{fig:seg_contrast}
\end{figure*}

\subsubsection{Quantitative Comparisons.}

To objectively evaluate MIND's performance, we conducted experiments on the Harvard Whole Brain dataset across three fusion tasks: CT-MRI, PET-MRI, and SPECT-MRI. As reported in Table \ref{tab:harvard_quant}, MIND demonstrates comprehensive superiority against eight state-of-the-art methods. Notably, it consistently achieves the highest scores in EN, MI, and SD across all modality pairs, proving its effectiveness in maximizing source information retention and maintaining high overall image contrast. When compared to state-of-the-art baselines, MIND yields substantial percentage improvements in metrics with significant gaps. For instance, it improves MI by 20.3\% over TextFusion \cite{CHENG2025102790} in the SPECT-MRI task, enhances SF by 13.1\% over BSAFusion \cite{Li_Su_Cai_Zhang_2025} in PET-MRI, and increases AG by 5.4\% over FILM \cite{ZHAO2024icml} in CT-MRI. Furthermore, MIND significantly minimizes performance variance, notably reducing the standard deviation of the SD metric in PET-MRI to just 2.907, a remarkable 71.3\% lower than FILM \cite{ZHAO2024icml}, ensuring highly robust and reliable fusion outcomes for clinical applications.

\subsubsection{Downstream Segmentation Task Validation.}
\begin{table}[!t]
\centering
\caption{Diffusion-based Model Efficiency Comparison on Harvard. Best results are highlighted as \colorbox{first}{\textbf{first}}, \colorbox{second}{second} and \colorbox{third}{third}. $\star$: using LoRA. Time: inference time per image. FLOPs: single-step flops.}
\vspace{-10pt}
\label{tab:efficiency_compare}
\begin{tabular}{lccc}
\toprule
\textbf{Methods} & \textbf{Params (M, $\downarrow$)} & \textbf{Time (s, $\downarrow$)} & \textbf{FLOPs (G, $\downarrow$)} \\
\midrule
DDFM \cite{Zhao_2023_ICCV}            & 552.814 & \cellcolor{third}{11.818} & \cellcolor{first}{\textbf{2293.76}} \\
Text-DiFuse \cite{NEURIPS2024_zhang}  & \cellcolor{third}{119.460} & \cellcolor{first}{\textbf{2.135}} & 18180.07 \\
DiTFuse$\star$ \cite{Li2025DiTFuse}   & \cellcolor{first}{\textbf{37.750}} & 13.370 & \cellcolor{second}{6469.60} \\
\textbf{MIND$\star$}                  & \cellcolor{second}{39.930} & \cellcolor{second}{7.851} & \cellcolor{third}{6860.69} \\
\bottomrule
\end{tabular}
\vspace{-10pt}
\end{table}

Table \ref{tab:brats_dice} and Fig. \ref{fig:seg_contrast} present the downstream segmentation results. Quantitative metrics reveal that single modalities exhibit inherent biases, such as MR-FLAIR excelling in ED but struggling with ET, and existing fusion methods like DDFM \cite{Zhao_2023_ICCV} and TextFusion \cite{CHENG2025102790} bridge this gap but often compromise on specific sub-regions. In contrast, MIND achieves the most balanced and comprehensive performance, securing the highest Dice scores for ED and ET, alongside a highly competitive NET score. Consequently, our method attains the optimal Mean Rank of 2.000 across all three sub-regions. Furthermore, visual comparisons with the top four performing baselines demonstrate that masks generated from competing methods like MACTFusion \cite{Xie2025MACTFusion} and TextFusion \cite{CHENG2025102790} often lack precision in complex internal boundaries. In contrast, our method produces masks with superior visual consistency to the Ground Truth, accurately delineating tumor sub-regions. This demonstrates the significant clinical value of our fused images for enhancing the accuracy of downstream medical image analysis.

\begin{table}[!t]
\centering
\caption{Dice for brain tumor segmentation on BraTS. Best results are highlighted as \colorbox{first}{\textbf{first}}, \colorbox{second}{second} and \colorbox{third}{third}. Mean Rank: the mean ranking across three metrics.}
\vspace{-10pt}
\label{tab:brats_dice}
\begin{tabular}{lcccc}
\toprule
\textbf{Methods} & \textbf{ED $\uparrow$} & \textbf{NET $\uparrow$} & \textbf{ET $\uparrow$} & \textbf{Mean Rank $\downarrow$} \\
\midrule
MR-FLAIR only            & \cellcolor{second}{0.785} & 0.482 & 0.450 & 7.333 \\
MR-T1CE only             & 0.701 & \cellcolor{second}{0.728} & \cellcolor{second}{0.588} & \cellcolor{third}{4.333} \\
DDFM \cite{Zhao_2023_ICCV}   & \cellcolor{third}{0.760} & \cellcolor{third}{0.712} & 0.552 & \cellcolor{second}{4.000} \\
Text-DiFuse \cite{NEURIPS2024_zhang}        & 0.735 & 0.625 & 0.528 & 6.333 \\
MACTFusion \cite{Xie2025MACTFusion}     & 0.757 & 0.705 & 0.568 & \cellcolor{third}{4.333} \\
BSAFusion \cite{Li_Su_Cai_Zhang_2025}      & 0.718 & 0.567 & 0.491 & 7.667 \\
TextFusion \cite{CHENG2025102790}         & 0.715 & \cellcolor{first}{\textbf{0.731}} & \cellcolor{third}{0.579} & \cellcolor{second}{4.000} \\
DiTFuse \cite{Li2025DiTFuse}    & 0.734 & 0.498 & 0.464 & 8.000 \\
SAFusion \cite{Li2026JBHI}        & 0.687 & 0.646 & 0.559 & 7.000 \\
\textbf{MIND}   & \cellcolor{first}{\textbf{0.786}} & 0.706 & \cellcolor{first}{\textbf{0.616}} & \cellcolor{first}{\textbf{2.000}} \\
\bottomrule
\end{tabular}
\vspace{-10pt}
\end{table}

\begin{table*}[!t]
\centering
\caption{Ablation study of our key contributions. Best results are highlighted as \colorbox{first}{\textbf{first}}, \colorbox{second}{second} and \colorbox{third}{third}. $\boldsymbol{\Phi}$: MLA, $\mathcal{L}_{sem}$: Medical Semantic Consistency Loss, $\boldsymbol{\gamma}$: Timestep-Truncation Mechanism. CLIP: the cosine similarity between the fused images and fusion texts.}
\vspace{-10pt}
\label{tab:ablation_components}
\setlength{\tabcolsep}{3pt}
\resizebox{\linewidth}{!}{
\begin{tabular}{ccc|cccccc|cccccc|cccccc}
\toprule
\multicolumn{3}{c|}{\textbf{Method}} & \multicolumn{6}{c|}{\textbf{Harvard CT-MRI}} & \multicolumn{6}{c|}{\textbf{Harvard PET-MRI}} & \multicolumn{6}{c}{\textbf{Harvard SPECT-MRI}} \\
\cmidrule(lr){1-3} \cmidrule(lr){4-9} \cmidrule(lr){10-15} \cmidrule(lr){16-21}
$\boldsymbol{\Phi}$ & $\mathcal{L}_{sem}$ & $\boldsymbol{\gamma}$ & 
\textbf{EN} & \textbf{AG} & \textbf{SF} & \textbf{VIF} & \textbf{Qabf} & \textbf{CLIP} &
\textbf{EN} & \textbf{AG} & \textbf{SF} & \textbf{VIF} & \textbf{Qabf} & \textbf{CLIP} &
\textbf{EN} & \textbf{AG} & \textbf{SF} & \textbf{VIF} & \textbf{Qabf} & \textbf{CLIP} \\
\midrule
& & & 3.267 & 7.233 & \cellcolor{third}{26.253} & 0.362 & 0.348 & 0.232 & 4.323 & 7.237 & 32.763 & 0.618 & 0.34 & 0.367 & 4.312 & 6.823 & 21.173 & 0.631 & 0.321 & 0.343 \\
\checkmark & & & 4.128 & 7.317 & 25.472 & \cellcolor{third}{0.403} & \cellcolor{second}{0.462} & 0.226 & \cellcolor{second}{5.028} & \cellcolor{third}{7.653} & 33.012 & \cellcolor{second}{0.732} & \cellcolor{third}{0.481} & 0.361 & \cellcolor{third}{5.138} & \cellcolor{first}{\textbf{7.012}} & \cellcolor{second}{22.601} & \cellcolor{third}{0.732} & \cellcolor{third}{0.477} & 0.335 \\
& \checkmark & \checkmark & 3.232 & \cellcolor{third}{7.963} & 25.431 & 0.398 & 0.374 & \cellcolor{first}{\textbf{0.321}} & 4.281 & 6.974 & \cellcolor{second}{35.136} & 0.683 & 0.362 & \cellcolor{second}{0.389} & 4.453 & 6.732 & 20.139 & 0.695 & 0.332 & \cellcolor{second}{0.472} \\
\checkmark & \checkmark & & \cellcolor{second}{5.032} & \cellcolor{second}{8.98} & \cellcolor{second}{32.05} & \cellcolor{first}{\textbf{0.571}} & \cellcolor{third}{0.412} & \cellcolor{third}{0.261} & \cellcolor{third}{4.972} & \cellcolor{second}{10.306} & \cellcolor{third}{33.819} & \cellcolor{third}{0.713} & \cellcolor{second}{0.663} & \cellcolor{first}{\textbf{0.393}} & \cellcolor{second}{5.895} & \cellcolor{third}{6.97} & \cellcolor{third}{21.938} & \cellcolor{first}{\textbf{0.886}} & \cellcolor{second}{0.615} & \cellcolor{third}{0.415} \\
\checkmark & \checkmark & \checkmark & \cellcolor{first}{\textbf{5.802}} & \cellcolor{first}{\textbf{10.111}} & \cellcolor{first}{\textbf{41.25}} & \cellcolor{second}{0.569} & \cellcolor{first}{\textbf{0.672}} & \cellcolor{second}{0.311} & \cellcolor{first}{\textbf{6.238}} & \cellcolor{first}{\textbf{11.893}} & \cellcolor{first}{\textbf{42.302}} & \cellcolor{first}{\textbf{0.749}} & \cellcolor{first}{\textbf{0.75}} & \cellcolor{third}{0.387} & \cellcolor{first}{\textbf{6.236}} & \cellcolor{second}{6.974} & \cellcolor{first}{\textbf{23.316}} & \cellcolor{second}{0.877} & \cellcolor{first}{\textbf{0.792}} & \cellcolor{first}{\textbf{0.473}} \\
\bottomrule
\end{tabular}
}
\end{table*}

\subsection{Computational Efficiency Analysis.}

We compare the computational efficiency of MIND against other diffusion-based methods. Table \ref{tab:efficiency_compare} reports three key metrics: model parameters (Params), inference time per image (Time), and single-step denoising floating-point operations (FLOPs). Both MIND and DiTFuse \cite{Li2025DiTFuse} use pre-trained DiT networks and are fine-tuned with LoRA. MIND requires only 39.930 M parameters and achieves an inference time of 7.851 s, making it the second most efficient method evaluated. While Text-DiFuse \cite{NEURIPS2024_zhang} and DiTFuse \cite{Li2025DiTFuse} are slightly more efficient in certain metrics, they perform noticeably worse in downstream modality fusion and segmentation tasks, which highlights the challenge of reducing computational overhead without sacrificing image quality. However, MIND maintains a competitive parameter count and inference speed while achieving state-of-the-art results in modality fusion and segmentation, offering a stronger trade-off between multimodal representation and computational cost.

\begin{figure}[thb]
\centering
\includegraphics[width=1.0\linewidth]{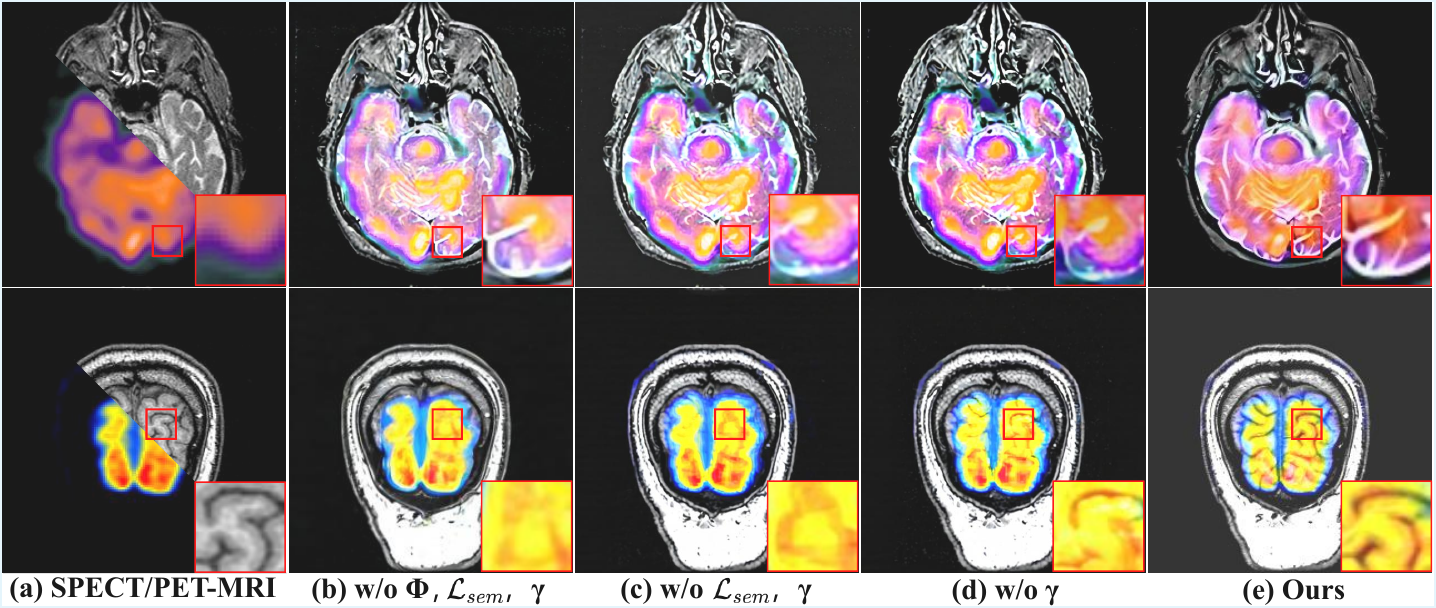}
\vspace{-20pt}
\caption{Ablation results on Harvard. Our complete MIND achieves the best visual quality, striking an optimal balance between pathological textures and metabolic hotspots.}
\vspace{-10pt}
\label{fig:ablation}
\end{figure}

\subsection{Ablation and Parameter Study.}
\subsubsection{Component Ablation.}
To validate the effectiveness of \textbf{MIND}, we conducted extensive ablation studies on Harvard. Both qualitative and quantitative results demonstrate that our complete MIND framework achieves optimal performance across most metrics. As illustrated in Fig. \ref{fig:ablation} (b) and (c), the incorporation of the MLA significantly enhances the preservation of source image information within the fused images. Quantitatively, Table \ref{tab:ablation_components} demonstrates that compared to the baseline without MLA, our method achieves an EN improvement of 46.76\% and 40.04\% on PET-MRI and SPECT-MRI subsets, respectively. Furthermore, as depicted in Fig. \ref{fig:ablation} (c) and (d), the Medical Semantic Consistency Loss effectively reinforces brain textures guided by the fusion intention, yielding results that are more semantically aligned with the target intention. Specifically, the CLIP score improves by 15.49\%, 8.86\%, and 23.88\% across the three subsets. However, this is accompanied by a degradation in VIF for PET-MRI and AG/SF for SPECT-MRI. This reveals an inherent conflict between semantic alignment and image fidelity: while the global semantic loss improves CLIP scores, it degrades visual quality. Thus, a well-tuned timestep-truncation threshold is essential to balance them. Notably, comparing the results in Fig. \ref{fig:ablation} (d) and (e) reveals that imposing this loss throughout the entire global reconstruction process leads to the loss of crucial texture details. Introducing the Timestep-Truncation Mechanism on PET-MRI yields a 20.39\% increase in SF, firmly validating the effectiveness of this mechanism in protecting early manifold generation. We thoroughly discuss the specific details of this mechanism in section \ref{sec:parameter} and Supplementary Material.

\begin{figure}[thb]
\centering
\includegraphics[width=1.0\linewidth]{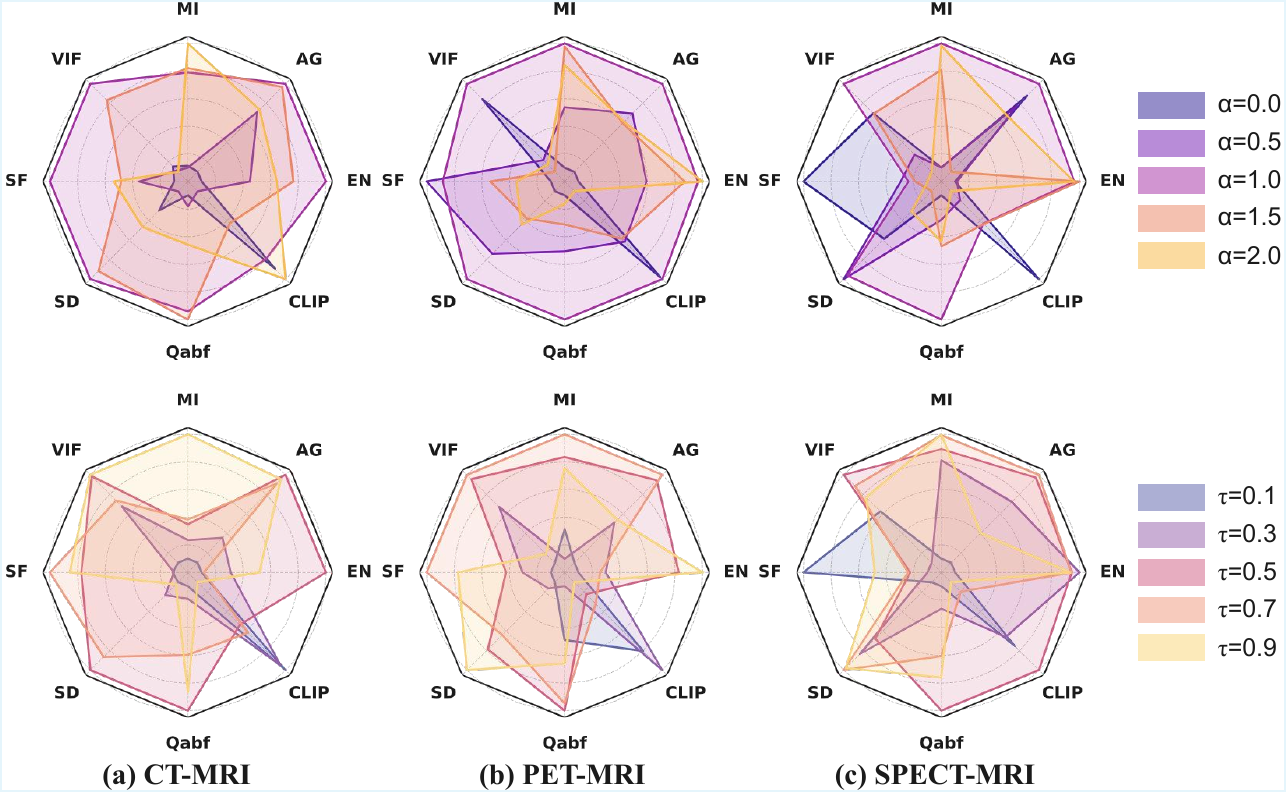}
\vspace{-20pt}
\caption{Hyperparameter analysis on Harvard. The metrics are normalized to [0,1] to show differences.}
\vspace{-10pt}
\label{fig_radar}
\end{figure}

\subsubsection{Hyperparameters Analysis.}
\label{sec:parameter}
We evaluate the adapter scale $\alpha$ and timestep-truncation threshold $\tau$ using normalized radar charts. Results indicate that the configuration of $\alpha=1.0$ and $\tau=0.5$ achieves a promising balance between visual quality and semantic alignment. As shown in Fig. \ref{fig_radar}, $\alpha$ modulates the injection intensity of source image features. Setting $\alpha=1.0$ yields the maximal and most balanced polygon area. A smaller scale of 0.5 provides insufficient prior compensation, causing a decline in MI and AG. Conversely, a larger scale ($\alpha=1.5$) over-amplifies the adapter feature, which disrupts the generative distribution of the pre-trained model and weaken visual quality. Furthermore, $\tau$ determines the intervention timing of semantic guidance. Lower thresholds ($\tau=0.1$) force premature semantic intervention, which severely degrades image fidelity metrics SF and VIF. Higher thresholds ($\tau=0.9$) preserve physical structures but fail to inject semantics effectively, leading to a significant drop in CLIP scores. Establishing $\tau=0.5$ successfully protects the physical skeleton during the early high-noise stage without semantic interference, and precisely leads diagnostic semantics during the later low-noise stage.

\section{Conclusion}
In this paper, we present MIND, a pioneering Multimodal Intent-Driven Network via DiTs that advances medical image fusion from rigid pixel-level rules to intelligent, clinical intent-driven generation. To overcome the spatial limitations of sequence modeling, our MLA explicitly injects 2D spatial features to preserve source features. Furthermore, our medical semantic consistency loss harmonizes physical fidelity with diagnostic intent by strictly isolating early-stage manifold generation from late-stage semantic injection. Extensive experiments demonstrate that MIND achieves state-of-the-art visual quality and significantly boosts downstream brain tumor segmentation accuracy. Although reliance on large vision-language models introduces computational overhead and rare-modality generalization requires further validation, future research on lightweight architectures will accelerate inference, ultimately advancing intent-driven clinical decision support systems.

\section{Acknowledgement}
This research was partially supported by National Natural Science Foundation of China under Grant No. U25D902, "Pioneer" and "Leading Goose" R\&D Program of Zhejiang under Grant No. 2024SSYS0026, and the Transvascular Implantation Devices Research Institute (TIDRI) under Grant No. KY052025003.

\clearpage
\appendix
\input{appendix}

\bibliographystyle{ACM-Reference-Format}
\bibliography{main} 

\end{document}

%% file: appendix.tex
\section*{Appendix}
\addcontentsline{toc}{section}{Appendix}
\vspace{1em}

\section{Details of our Training Dataset}
Our primary training dataset is sourced from Harvard and BraTS 2017 \cite{brats_dataset}. It consists of 160 CT-MRI pairs, 245 SPECT-MRI pairs, 333 SPECT-MRI pairs and 387 FLAIR-T1CE pairs. We utilize prompts enriched with medical prior knowledge to guide BioMedGPT in generating intent-driven fusion text descriptions for different multimodal image pairs. Specifically, for CT-MRI pairs, we direct the model to focus on the anatomical structures of high-density tissues in CT and the rich soft-tissue details and high-contrast features in MRI. For PET-MRI pairs, we guide the model to extract the metabolic and functional activity features from PET and integrate them with the precise anatomical localization of MRI. For SPECT-MRI pairs, we employ prompts to fuse the blood perfusion information captured by SPECT with the morphological references provided by MRI. Furthermore, the distinct tags within the fusion prefixes strengthen the model's ability to discriminate between various modalities. Notably, to mitigate the challenges of data scarcity and the lack of absolute Ground Truth (GT) for multimodal fusion, we extend the complementary degradation training paradigm \cite{Li2025DiTFuse} to medical imaging scenarios. Specifically, given a high-quality, single-modality MRI image ($I_{clean}$), we generate two distinct degraded versions ($I_{deg1}$ and $I_{deg2}$) by injecting independent noise profiles and artifacts consistent with medical imaging distributions. Crucially, these degradations are designed to be spatially and semantically complementary. By inputting the degraded pair $\{I_{deg1}, I_{deg2}\}$, we construct a self-supervised training configuration where the original $I_{clean}$ serves as the genuine, natural GT. This enables the model to explicitly learn how to extract and organically fuse complementary features from corrupted modalities.

\begin{figure}[thb]
\centering
\includegraphics[width=1.0\linewidth]{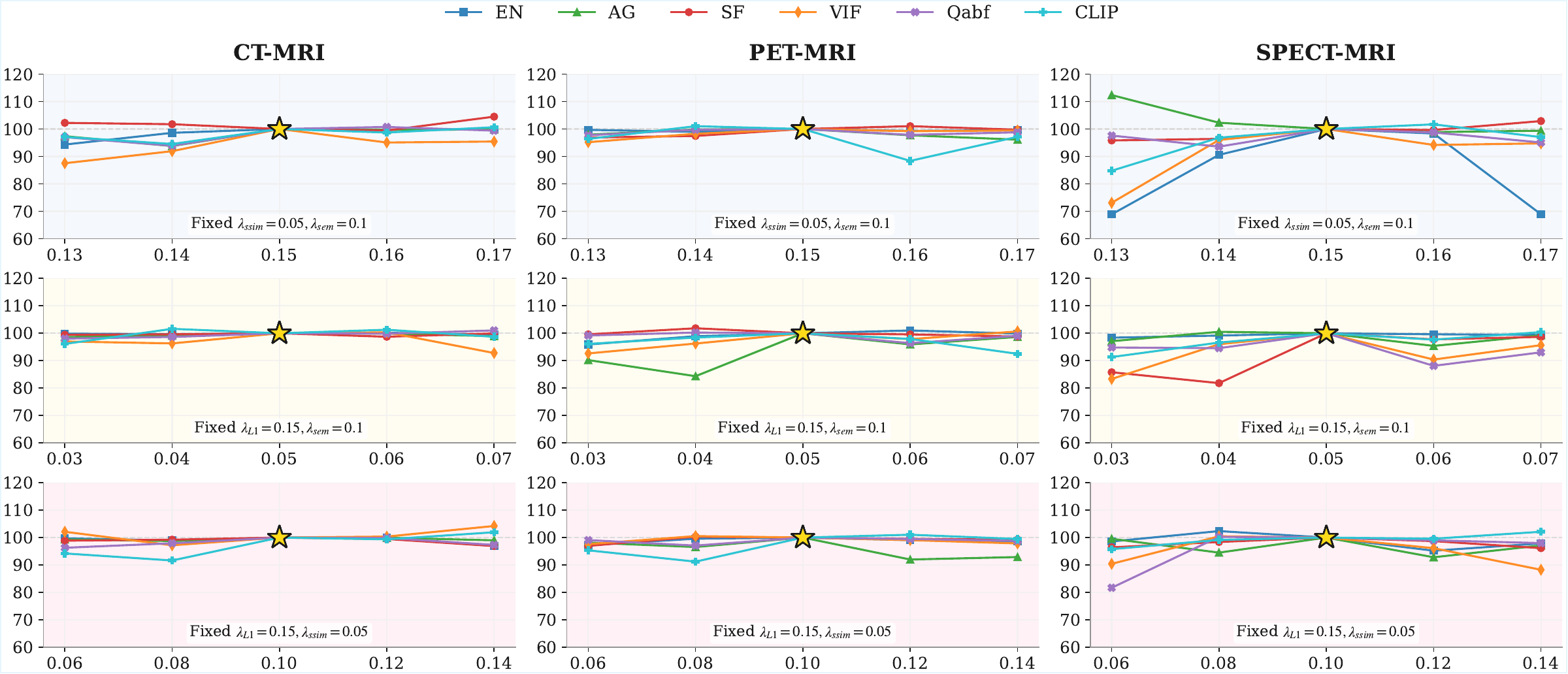}
\vspace{-20pt}
\caption{Sensitivity analysis of loss weights on Harvard. Metrics are normalized with the optimal configuration set to 100\%. The star highlights the optimal weight combination adopted in our model.}
\label{fig:loss_weights}
\end{figure}
\section{Hyperparamter Analysis}
We conducted a detailed sensitivity analysis on the weights of the loss functions. As shown in Fig. \ref{fig:loss_weights}, we froze two of the weights to investigate the optimal value of the remaining one. When freezing $\lambda_{ssim}=0.05$ and $\lambda_{sem}=0.10$, increasing the $\lambda_{L1}$ weight from 0.13 to 0.15 achieves significant gains in image quality. Specifically, on the CT-MRI dataset, the visual information fidelity VIF increases by 14.3\% and the entropy EN improves by 6.0\%. Furthermore, increasing the $\lambda_{ssim}$ weight from 0.03 to 0.05 significantly improves the structural preservation and spatial frequency of the fused images, where VIF and SF on the SPECT-MRI substantially increase by 20.0\% and 16.6\% respectively. However, further increasing this weight to 0.07 causes a significant drop in edge information retention Qabf, marking a decrease of 6.9\% on the SPECT-MRI, which leads to an imbalance in overall image quality. Additionally, setting the $\lambda_{sem}$ weight as low as 0.06 degrades the model's performance in semantic consistency, resulting in an average drop of 5\% in the CLIP score across all datasets. Regarding the threshold $\theta$ in $\mathcal{L}_{sem}$, when the image-text cosine similarity exceeds this threshold, we set this loss term to 0, preventing texture distortion caused by over-alignment when correct texts are provided. Inspired by existing advanced multimodal alignment methods \cite{Xiang25JBHI} and combined with empirical settings, we set the threshold to 0.85 to ensure sufficient semantic alignment while effectively preventing model overfitting.

To balance the visual quality of the fused images and the interactive performance of the model, we ultimately selected $\lambda_{L1}=0.15$, $\lambda_{ssim}=0.05$, and $\lambda_{sem}=0.10$ as the optimal weights for the loss terms.

\begin{figure*}[thb]
\centering
\includegraphics[width=1.0\linewidth]{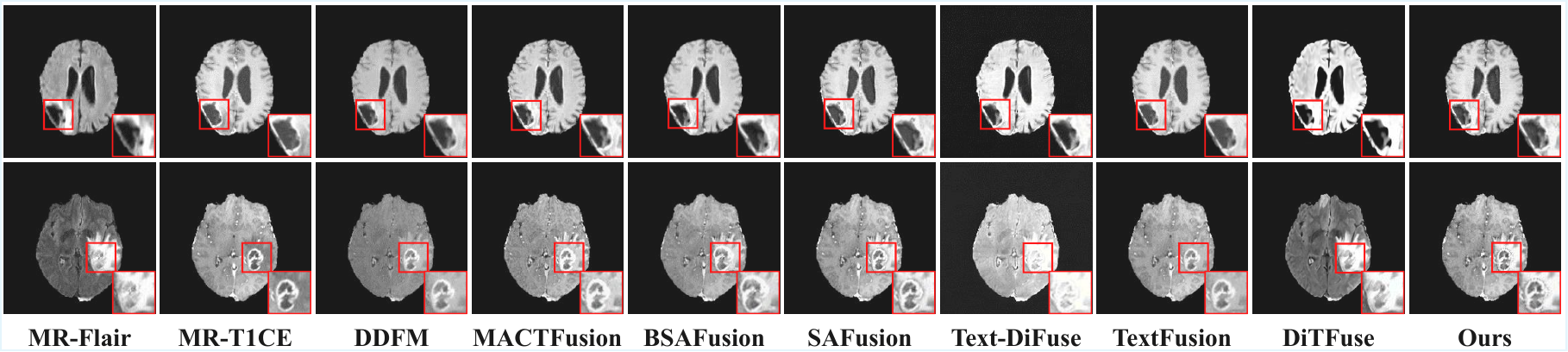}
\vspace{-20pt}
\caption{Extra FLAIR-T1CE fusion results on BraTS 2017 \cite{brats_dataset}. Red boxes highlight fine-grained local details of tumors.}
\vspace{-10pt}
\label{fig:brats_contrast}
\end{figure*}

\begin{figure*}[thb]
\centering
\includegraphics[width=1.0\linewidth]{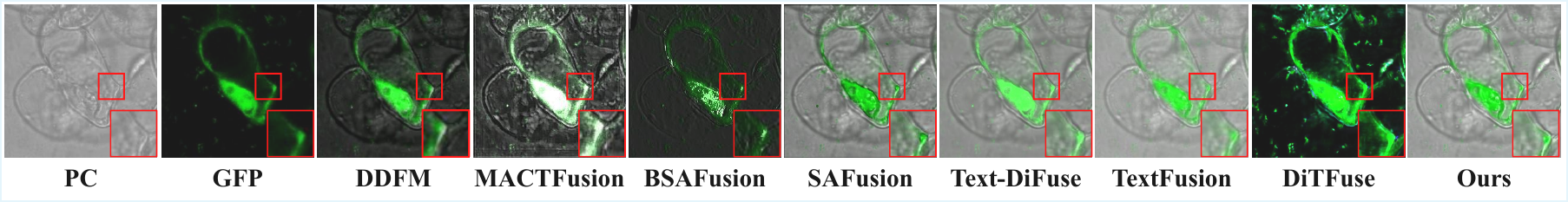}
\vspace{-20pt}
\caption{Extra GFP-PC fusion results on GFP \cite{koroleva2005high}. Red boxes highlight fine-grained local details.}
\vspace{-10pt}
\label{fig:gfp_contrast}
\end{figure*}

\begin{table*}[!t]
\centering
\caption{Quantitative comparison of MIND with eight state-of-the-art methods on BraTS 2017 \cite{brats_dataset} and GFP \cite{koroleva2005high}. Best results are highlighted as \colorbox{first}{\textbf{first}}, \colorbox{second}{second} and \colorbox{third}{third}.}
\vspace{-10pt}
\label{tab:quant_comparison}
\footnotesize
\setlength{\tabcolsep}{3pt}  
\resizebox{\linewidth}{!}{
\begin{tabular}{lllcccccccc}
\toprule
\textbf{Dataset} & \textbf{Method} & \textbf{Source} & \textbf{EN} & \textbf{AG} & \textbf{MI} & \textbf{VIF} & \textbf{SF} & \textbf{SD} & \textbf{Qabf} & \textbf{MS\_SSIM} \\
\midrule

\multirow{9}{*}{\textbf{FLAIR-T1CE}}
    & DDFM \cite{Zhao_2023_ICCV}           & ICCV$_{23}$          & 2.772±0.273 & 2.258±0.378 & 2.775±0.218 & 0.807±0.105 & 12.156±1.949 & 68.529±9.097 & 0.539±0.069 & 0.832±0.049 \\
    & Text-DiFuse \cite{NEURIPS2024_zhang}   & NeurIPS$_{24}$       & 2.586±0.338 & 3.221±0.567 & 2.612±0.227 & 0.719±0.084 & \cellcolor{second}{18.418±2.432} & \cellcolor{second}{89.519±7.812} & 0.488±0.055 & 0.516±0.043 \\
    & FILM \cite{ZHAO2024icml}            & ICML$_{24}$          & 2.792±0.427 & 3.023±0.371 & \cellcolor{third}{2.813±0.219} & 0.712±0.129 & 14.525±2.584 & 76.752±9.117 & 0.512±0.071 & 0.798±0.062 \\
    & MACTFusion \cite{Xie2025MACTFusion}  & JBHI$_{25}$          & 2.833±0.283 & \cellcolor{second}{3.410±0.579} & 2.713±0.213 & 0.795±0.099 & 16.722±2.349 & 76.567±8.901 & \cellcolor{second}{0.628±0.044} & \cellcolor{second}{0.891±0.042} \\
    & BSAFusion \cite{Li_Su_Cai_Zhang_2025}     & AAAI$_{25}$          & \cellcolor{third}{2.872±0.282} & 2.805±0.498 & 2.668±0.224 & 0.653±0.113 & 13.755±2.044 & 75.770±8.880 & 0.482±0.076 & \cellcolor{first}{\textbf{0.896±0.043}} \\
    & TextFusion \cite{CHENG2025102790}    & INFFUS$_{25}$        & 2.841±0.284 & 2.658±0.497 & \cellcolor{first}{\textbf{3.139±0.278}} & \cellcolor{first}{\textbf{0.967±0.112}} & 14.356±2.163 & 77.729±10.569 & \cellcolor{third}{0.617±0.046} & 0.875±0.048 \\
    & DiTFuse \cite{Li2025DiTFuse}        & TPAMI$_{25}$         & \cellcolor{second}{2.926±0.342} & 2.915±0.647 & 2.655±0.239 & 0.609±0.070 & 15.499±3.939 & 72.530±12.624 & 0.581±0.055 & 0.805±0.096 \\
    & SAFusion \cite{Li2026JBHI}       & JBHI$_{26}$          & 2.851±0.288 & \cellcolor{third}{3.362±0.586} & \cellcolor{second}{3.077±0.307} & \cellcolor{third}{0.871±0.101} & \cellcolor{third}{17.124±2.438} & \cellcolor{third}{79.703±9.608} & \cellcolor{first}{\textbf{0.637±0.032}} & 0.868±0.047 \\
    & \textbf{MIND}     & \textbf{Ours}                 & \cellcolor{first}{\textbf{4.457±0.332}} & \cellcolor{first}{\textbf{3.424±0.501}} & 2.522±0.213 & \cellcolor{second}{0.887±0.063} & \cellcolor{first}{\textbf{18.506±2.506}} & \cellcolor{first}{\textbf{90.557±9.871}} & 0.597±0.043 & \cellcolor{third}{0.876±0.049} \\
\midrule

\multirow{9}{*}{\textbf{GFP-PC}}
    & DDFM \cite{Zhao_2023_ICCV}           & ICCV$_{23}$          & 6.764±0.306 & 4.265±0.898 & 3.068±0.692 & 0.882±0.141 & 10.385±2.311 & 31.658±4.058 & 0.575±0.126 & \cellcolor{second}{0.485±0.021} \\
    & FILM \cite{ZHAO2024icml}            & ICML$_{24}$          & 6.562±0.374 & 4.528±1.077 & 3.108±0.467 & 0.721±0.137 & 10.792±3.292 & 35.485±8.231 & 0.439±0.153 & 0.429±0.051 \\
    & Text-DiFuse \cite{NEURIPS2024_zhang}   & NeurIPS$_{24}$       & 6.965±0.329 & 4.123±0.934 & 2.735±0.279 & 0.805±0.102 & 9.075±2.722 & 32.751±5.680 & 0.533±0.070 & \cellcolor{third}{0.461±0.028} \\
    & MACTFusion \cite{Xie2025MACTFusion}  & JBHI$_{25}$          & \cellcolor{second}{7.196±0.263} & \cellcolor{third}{5.676±3.495} & 1.847±0.278 & 0.679±0.161 & \cellcolor{first}{\textbf{28.842±10.601}} & \cellcolor{second}{56.414±6.607} & 0.412±0.061 & 0.236±0.085 \\
    & BSAFusion \cite{Li_Su_Cai_Zhang_2025}     & AAAI$_{25}$          & 5.472±1.133 & 4.325±1.882 & 1.214±0.425 & 0.242±0.149 & \cellcolor{second}{20.684±6.502} & 29.259±8.651 & 0.531±0.167 & 0.206±0.130 \\
    & TextFusion  \cite{CHENG2025102790}    & INFFUS$_{25}$        & 6.350±0.484 & 3.367±1.273 & \cellcolor{first}{\textbf{4.453±0.393}} & \cellcolor{second}{0.940±0.057} & 7.793±3.097 & 22.078±6.651 & \cellcolor{second}{0.708±0.125} & \cellcolor{first}{\textbf{0.496±0.006}} \\
    & DiTFuse \cite{Li2025DiTFuse}        & TPAMI$_{25}$         & \cellcolor{third}{7.094±0.468} & \cellcolor{second}{5.849±2.061} & 2.561±0.367 & 0.699±0.097 & 12.881±4.687 & \cellcolor{third}{38.778±11.283} & 0.349±0.070 & 0.417±0.033 \\
    & SAFusion \cite{Li2026JBHI}       & JBHI$_{26}$          & 7.017±0.273 & 5.197±0.948 & \cellcolor{third}{3.906±0.363} & \cellcolor{first}{\textbf{0.942±0.102}} & 10.855±2.107 & 34.698±6.336 & \cellcolor{third}{0.604±0.105} & 0.447±0.036 \\
    & \textbf{MIND}     & \textbf{Ours}                 & \cellcolor{first}{\textbf{7.296±0.722}} & \cellcolor{first}{\textbf{5.926±2.261}} & \cellcolor{second}{4.239±0.784} & \cellcolor{third}{0.928±0.283} & \cellcolor{third}{18.842±10.601} & \cellcolor{first}{\textbf{58.221±12.143}} & \cellcolor{first}{\textbf{0.769±0.125}} & 0.453±0.129 \\
\bottomrule
\end{tabular}
}
\end{table*}

\section{Extra Details of Comparative Experiment}

Fig. \ref{fig:brats_contrast} and \ref{fig:gfp_contrast} show qualitative fusion results across different modalities. On the BraTS dataset \cite{brats_dataset}, methods such as DDFM \cite{Zhao_2023_ICCV}, MACTFusion 
\cite{Xie2025MACTFusion}, and TextFusion \cite{CHENG2025102790} mainly preserve the core tumor from the T1CE modality but often fail to retain the surrounding edema, whereas Text-DiFuse \cite{NEURIPS2024_zhang} and DiTFuse \cite{Li2025DiTFuse} show the opposite behavior. In comparison, MIND achieves a more balanced integration, preserving both clear tumor boundaries and edema regions from the FLAIR modality. This ability to capture complementary features is also observed in the external GFP-PC validation, where MIND consistently combines the structural information of the PC modality with the fluorescence patterns of the GFP modality, yielding stable visual results across different imaging scenarios.

Table \ref{tab:quant_comparison} reports the quantitative results, where MIND performs competitively on both datasets. On FLAIR-T1CE, MIND achieves the highest scores in EN, AG, SF, and SD, surpassing DiTFuse \cite{Li2025DiTFuse} by 52.3\% in EN and improving SD by 1.16\% over Text-DiFuse \cite{NEURIPS2024_zhang}. A similar pattern is observed in the GFP-PC validation, where MIND leads in EN, AG, SD, and Qabf. While MACTFusion \cite{Xie2025MACTFusion} presents an anomalously high SF score in this task, it suffers a severe degradation in MS\_SSIM. In contrast, MIND maintains a more balanced performance, avoiding severe structural degradation on unseen data while providing steady improvements in overall fusion quality.

\begin{table*}[!t]
\centering
\caption{Model robustness study on different instructions. Best results are highlighted as \colorbox {first}{\textbf {first}}, \colorbox {second}{second} and \colorbox {third}{third}. CLIP: the cosine similarity between the fused images and fusion texts. Original: using their original style or process-driven fusion texts for each methods. Wrong: using wrong fusion texts. Intent-Driven: using Intent-Driven fusion texts to describe details of the desired fusion results.}
\vspace{-10pt}
\label{tab:ablation_instruction}
\setlength{\tabcolsep}{3pt}
\resizebox{\linewidth}{!}{
\begin{tabular}{cc|cccccc|cccccc|cccccc}
\toprule
\multirow{2}{*}{\textbf{Method}} & \multirow{2}{*}{\textbf{Instruction}} & \multicolumn{6}{c|}{\textbf{CT-MRI}} & \multicolumn{6}{c|}{\textbf{PET-MRI}} & \multicolumn{6}{c}{\textbf{SPECT-MRI}} \\
\cmidrule(lr){3-8} \cmidrule(lr){9-14} \cmidrule(lr){15-20}
& & \textbf{EN} & \textbf{AG} & \textbf{SF} & \textbf{VIF} & \textbf{Qabf} & \textbf{CLIP} &
\textbf{EN} & \textbf{AG} & \textbf{SF} & \textbf{VIF} & \textbf{Qabf} & \textbf{CLIP} &
\textbf{EN} & \textbf{AG} & \textbf{SF} & \textbf{VIF} & \textbf{Qabf} & \textbf{CLIP} \\
\midrule
\multirow{3}{*}{FILM} 
& Original & \cellcolor{second}{3.907} & \cellcolor{first}{\textbf{9.593}} & \cellcolor{second}{42.957} & \cellcolor{second}{0.403} & \cellcolor{first}{\textbf{0.644}} & \cellcolor{third}{0.328} & \cellcolor{second}{3.437} & \cellcolor{second}{8.121} & \cellcolor{first}{\textbf{31.297}} & \cellcolor{first}{\textbf{0.793}} & \cellcolor{first}{\textbf{0.784}} & \cellcolor{first}{\textbf{0.361}} & \cellcolor{first}{\textbf{3.776}} & \cellcolor{second}{5.728} & \cellcolor{second}{22.687} & \cellcolor{second}{0.754} & \cellcolor{first}{\textbf{0.671}} & \cellcolor{second}{0.412} \\
& Wrong & \cellcolor{third}{3.671} & \cellcolor{third}{8.472} & \cellcolor{third}{34.831} & \cellcolor{third}{0.389} & \cellcolor{second}{0.583} & \cellcolor{second}{0.331} & \cellcolor{third}{2.971} & \cellcolor{third}{6.915} & \cellcolor{third}{27.773} & \cellcolor{third}{0.501} & \cellcolor{third}{0.618} & \cellcolor{third}{0.237} & \cellcolor{second}{3.314} & \cellcolor{third}{5.337} & \cellcolor{third}{19.873} & \cellcolor{third}{0.631} & \cellcolor{third}{0.543} & \cellcolor{third}{0.337} \\
& Intent-Driven & \cellcolor{first}{\textbf{3.913}} & \cellcolor{second}{9.438} & \cellcolor{first}{\textbf{43.489}} & \cellcolor{first}{\textbf{0.412}} & \cellcolor{third}{0.547} & \cellcolor{first}{\textbf{0.337}} & \cellcolor{first}{\textbf{3.871}} & \cellcolor{first}{\textbf{8.193}} & \cellcolor{second}{29.984} & \cellcolor{second}{0.778} & \cellcolor{second}{0.647} & \cellcolor{second}{0.313} & \cellcolor{third}{3.137} & \cellcolor{first}{\textbf{5.810}} & \cellcolor{first}{\textbf{25.779}} & \cellcolor{first}{\textbf{0.773}} & \cellcolor{second}{0.634} & \cellcolor{first}{\textbf{0.437}} \\
\midrule
\multirow{3}{*}{TextFusion} 
& Original & \cellcolor{second}{4.775} & \cellcolor{second}{5.020} & \cellcolor{first}{\textbf{22.424}} & \cellcolor{second}{0.436} & \cellcolor{second}{0.380} & \cellcolor{second}{0.301} & \cellcolor{second}{5.770} & \cellcolor{first}{\textbf{9.983}} & \cellcolor{second}{32.018} & \cellcolor{second}{0.633} & \cellcolor{second}{0.696} & \cellcolor{second}{0.283} & \cellcolor{second}{5.032} & \cellcolor{second}{6.103} & \cellcolor{second}{20.240} & \cellcolor{second}{0.672} & \cellcolor{second}{0.677} & \cellcolor{second}{0.399} \\
& Wrong & \cellcolor{third}{4.765} & \cellcolor{third}{4.547} & \cellcolor{third}{20.704} & \cellcolor{third}{0.384} & \cellcolor{third}{0.338} & \cellcolor{third}{0.215} & \cellcolor{third}{5.764} & \cellcolor{third}{9.565} & \cellcolor{third}{29.856} & \cellcolor{third}{0.599} & \cellcolor{third}{0.605} & \cellcolor{third}{0.251} & \cellcolor{third}{5.025} & \cellcolor{third}{6.088} & \cellcolor{third}{20.111} & \cellcolor{third}{0.665} & \cellcolor{third}{0.675} & \cellcolor{third}{0.369} \\
& Intent-Driven & \cellcolor{first}{\textbf{4.858}} & \cellcolor{first}{\textbf{5.205}} & \cellcolor{second}{22.107} & \cellcolor{first}{\textbf{0.445}} & \cellcolor{first}{\textbf{0.384}} & \cellcolor{first}{\textbf{0.342}} & \cellcolor{first}{\textbf{5.803}} & \cellcolor{second}{9.740} & \cellcolor{first}{\textbf{32.156}} & \cellcolor{first}{\textbf{0.644}} & \cellcolor{first}{\textbf{0.714}} & \cellcolor{first}{\textbf{0.298}} & \cellcolor{first}{\textbf{5.059}} & \cellcolor{first}{\textbf{6.401}} & \cellcolor{first}{\textbf{20.566}} & \cellcolor{first}{\textbf{0.684}} & \cellcolor{first}{\textbf{0.693}} & \cellcolor{first}{\textbf{0.404}} \\
\midrule
\multirow{3}{*}{DiTFuse} 
& Original & \cellcolor{second}{3.884} & \cellcolor{second}{8.979} & \cellcolor{first}{\textbf{32.050}} & \cellcolor{first}{\textbf{0.412}} & \cellcolor{first}{\textbf{0.471}} & \cellcolor{second}{0.293} & \cellcolor{first}{\textbf{4.868}} & \cellcolor{first}{\textbf{12.423}} & \cellcolor{second}{33.819} & \cellcolor{second}{0.557} & \cellcolor{first}{\textbf{0.640}} & \cellcolor{second}{0.371} & \cellcolor{second}{4.275} & \cellcolor{first}{\textbf{6.947}} & \cellcolor{second}{24.304} & \cellcolor{second}{0.570} & \cellcolor{second}{0.553} & \cellcolor{first}{\textbf{0.441}} \\
& Wrong & \cellcolor{third}{3.623} & \cellcolor{third}{6.747} & \cellcolor{second}{25.473} & \cellcolor{third}{0.307} & \cellcolor{third}{0.414} & \cellcolor{third}{0.271} & \cellcolor{third}{3.237} & \cellcolor{third}{10.881} & \cellcolor{third}{31.429} & \cellcolor{third}{0.423} & \cellcolor{third}{0.521} & \cellcolor{third}{0.274} & \cellcolor{third}{3.833} & \cellcolor{third}{5.764} & \cellcolor{first}{\textbf{24.312}} & \cellcolor{third}{0.413} & \cellcolor{third}{0.474} & \cellcolor{third}{0.365} \\
& Intent-Driven & \cellcolor{first}{\textbf{3.898}} & \cellcolor{first}{\textbf{8.988}} & \cellcolor{third}{24.793} & \cellcolor{second}{0.331} & \cellcolor{second}{0.427} & \cellcolor{first}{\textbf{0.303}} & \cellcolor{second}{4.734} & \cellcolor{second}{11.730} & \cellcolor{first}{\textbf{34.187}} & \cellcolor{first}{\textbf{0.631}} & \cellcolor{second}{0.544} & \cellcolor{first}{\textbf{0.383}} & \cellcolor{first}{\textbf{5.031}} & \cellcolor{second}{6.798} & \cellcolor{third}{22.518} & \cellcolor{first}{\textbf{0.584}} & \cellcolor{first}{\textbf{0.648}} & \cellcolor{second}{0.397} \\
\midrule
\multirow{3}{*}{\textbf{MIND}} 
& Original & \cellcolor{second}{5.475} & \cellcolor{second}{9.739} & \cellcolor{second}{35.008} & \cellcolor{second}{0.423} & \cellcolor{second}{0.588} & \cellcolor{third}{0.293} & \cellcolor{second}{5.973} & \cellcolor{first}{\textbf{11.978}} & \cellcolor{first}{\textbf{43.095}} & \cellcolor{second}{0.703} & \cellcolor{second}{0.637} & \cellcolor{second}{0.379} & \cellcolor{second}{6.118} & \cellcolor{second}{6.331} & \cellcolor{second}{21.832} & \cellcolor{second}{0.835} & \cellcolor{second}{0.693} & \cellcolor{third}{0.371} \\
& Wrong & \cellcolor{third}{4.352} & \cellcolor{third}{7.014} & \cellcolor{third}{26.423} & \cellcolor{third}{0.371} & \cellcolor{third}{0.324} & \cellcolor{first}{\textbf{0.327}} & \cellcolor{third}{5.475} & \cellcolor{third}{8.674} & \cellcolor{third}{29.994} & \cellcolor{third}{0.413} & \cellcolor{third}{0.552} & \cellcolor{third}{0.332} & \cellcolor{third}{4.136} & \cellcolor{third}{5.024} & \cellcolor{third}{19.774} & \cellcolor{third}{0.752} & \cellcolor{third}{0.529} & \cellcolor{second}{0.401} \\
& Intent-Driven & \cellcolor{first}{\textbf{5.802}} & \cellcolor{first}{\textbf{10.111}} & \cellcolor{first}{\textbf{41.250}} & \cellcolor{first}{\textbf{0.569}} & \cellcolor{first}{\textbf{0.672}} & \cellcolor{second}{0.311} & \cellcolor{first}{\textbf{6.238}} & \cellcolor{second}{11.893} & \cellcolor{second}{42.302} & \cellcolor{first}{\textbf{0.749}} & \cellcolor{first}{\textbf{0.750}} & \cellcolor{first}{\textbf{0.387}} & \cellcolor{first}{\textbf{6.236}} & \cellcolor{first}{\textbf{6.974}} & \cellcolor{first}{\textbf{23.316}} & \cellcolor{first}{\textbf{0.877}} & \cellcolor{first}{\textbf{0.792}} & \cellcolor{first}{\textbf{0.473}} \\
\bottomrule
\end{tabular}
}
\vspace{-10pt}
\end{table*}
\section{Model Rubustness Test}

The core of intent-driven fusion texts lies in utilizing specialized medical terminology to describe the pathological features of expected fused images. In contrast, prior fusion methods predominantly rely on general photographic descriptions to articulate source image features or the fusion process, termed process-driven. We compare the impact of different fusion texts on image quality across text-driven methods. Notably, we introduce mismatched texts that clearly contradict the source images to test model robustness. For example, in a SPECT-MRI fusion scene exhibiting prominent focal abnormal perfusion and brain lesions, we deliberately input a text describing "a completely healthy cerebral hemisphere with uniform metabolic distribution and no pathological structural deformities", directly conflicting with the actual visual evidence. 

As shown in Table \ref{tab:ablation_instruction}, our intent-driven prompts yield competitive performance gains across all evaluated methods. Specifically, for DiTFuse, the VIF on the PET-MRI dataset increases by 13.3\%. FILM and TextFusion also exhibit marginal improvements; for instance, FILM shows a slight increase in VIF on CT-MRI, while the EN of TextFusion on CT-MRI improves by 1.7\%. For the proposed MIND, intent-driven texts significantly boost the VIF by 34.5\% and 6.5\% on the CT-MRI and PET-MRI datasets respectively compared to original texts. Furthermore, MIND demonstrates exceptional robustness, suffering minimal performance degradation when perturbed by incorrect fusion texts. For example, regarding VIF on the SPECT-MRI dataset, DiTFuse and FILM experience steep declines of 27.5\% and 16.3\% respectively. Crucially, even when driven by incorrect texts, the EN of MIND on CT-MRI remains at 4.352, which still outperforms FILM and DiTFuse equipped with original texts by 11.4\% and 12.0\% respectively. MIND maintains high quality across core metrics such as EN and AG under erroneous text interference. This resilience stems from the delicate balance within our multi-objective loss function. The underlying flow matching and reconstruction losses establish a strict lower bound for visual quality. They act as robust anchors to prevent erroneous semantic guidance from catastrophically disrupting critical anatomical structures, thereby enabling the model to preserve fundamental medical characteristics from the source images as a reliable baseline.

\section{Allocation Strategy Analysis}
To fully investigate the effectiveness of linearly allocation strategy in MLA, Table \ref{tab:allocation_strategy} compares the differences in image quality and the corresponding downstream brain tumor segmentation performance under three alternative allocation strategies. For Strategy A (Reversed Allocation), we inject the low-resolution Scale-2 into the shallow layers (0-10) and the high-resolution Scale-0 into the deep layers (22-31). Since this reversed allocation contradicts the natural progression of Transformers in extracting features from spatial details to high-level semantics, it causes severe difficulties in feature integration. Consequently, its EN is only 2.863, and the VIF drops to 0.851, which directly hinders the fused images from presenting clear structural contrast between the lesions and normal brain tissues. For Strategy B (Uniform Injection), we make no hierarchical distinctions, concatenating features from all three scales (Scale-0, 1, and 2) and injecting them simultaneously into all Transformer layers. Although this approach achieves the second-best EN (2.888) and VIF (0.885) through aggressive feature stacking, severe scale confusion and feature redundancy result in the lowest Qabf and SF (0.512 and 17.074 respectively). Such degradation inevitably blurs lesion boundaries and texture details, significantly diminishing the diagnostic value of the medical images. For Strategy C (Single Scale), we extract only the highest-resolution Scale-0 and inject it across all 32 layers. Lacking multi-scale contextual guidance, this strategy retains some high-frequency details with a second-best Qabf of 0.541 but suffers from insufficient global information extraction, yielding an EN of merely 2.862. In contrast, our linear allocation strategy successfully demonstrates perfect adaptability to the network depth by aligning high-to-low resolution scales with the shallow-to-deep layers. It achieves the best performance across all core image fusion quality metrics, notably peaking at 4.457 for EN and 0.597 for Qabf. 

Building upon these superior fusion metrics, the downstream segmentation results further illuminate the critical link between image quality and practical clinical value. The enhanced visual fidelity achieved by our linear allocation strategy directly translates into robust diagnostic utility, yielding the highest mean DICE of 0.703 for brain tumor segmentation. Compared with the strategies A, B, and C, our method effectively preserves structural coherence and clear lesion boundaries. Ultimately, this distinct performance gap confirms that aligning multi-scale feature injection with the natural depth progression of the network is fundamental for reliable and accurate medical image analysis.

\begin{table}[t]
\centering
\caption{Quantitative comparison on BraTS 2017 \cite{brats_dataset} of different allocation strategies in MLA. Best results are highlighted as \colorbox {first}{\textbf {first}}, \colorbox {second}{second} and \colorbox {third}{third}. DICE (Mean): Mean dice score on 3 kinds of tumor segmentations.}
\vspace{-10pt}
\label{tab:allocation_strategy}
\footnotesize
\setlength{\tabcolsep}{3pt}
\resizebox{\linewidth}{!}{
\begin{tabular}{ccccccc}
\toprule
\textbf{Allocation Strategy} & \textbf{EN} & \textbf{AG} & \textbf{SF} & \textbf{VIF} & \textbf{Qabf} & \textbf{DICE (Mean)} \\
\midrule
Variant A (Reversed)           & \cellcolor{third}{2.863} & 3.361 & \cellcolor{second}{17.765} & 0.851 & \cellcolor{third}{0.533} & \cellcolor{second}{0.671} \\
Variant B (Uniform Injection)   & \cellcolor{second}{2.888} & \cellcolor{second}{3.418} & 17.074 & \cellcolor{second}{0.885} & 0.512 & 0.667 \\
Variant C (Single Scale)       & 2.862 & \cellcolor{third}{3.404} & \cellcolor{third}{17.471} & \cellcolor{third}{0.865} & \cellcolor{second}{0.541} & \cellcolor{third}{0.669} \\
\textbf{Ours (Linear Allocation)} & \cellcolor{first}{\textbf{4.457}} & \cellcolor{first}{\textbf{3.424}} & \cellcolor{first}{\textbf{18.506}} & \cellcolor{first}{\textbf{0.887}} & \cellcolor{first}{\textbf{0.597}} & \cellcolor{first}{\textbf{0.703}} \\
\bottomrule
\end{tabular}
}
\vspace{-10pt}
\end{table}

\begin{figure*}[thb]
\centering
\includegraphics[width=1.0\linewidth]{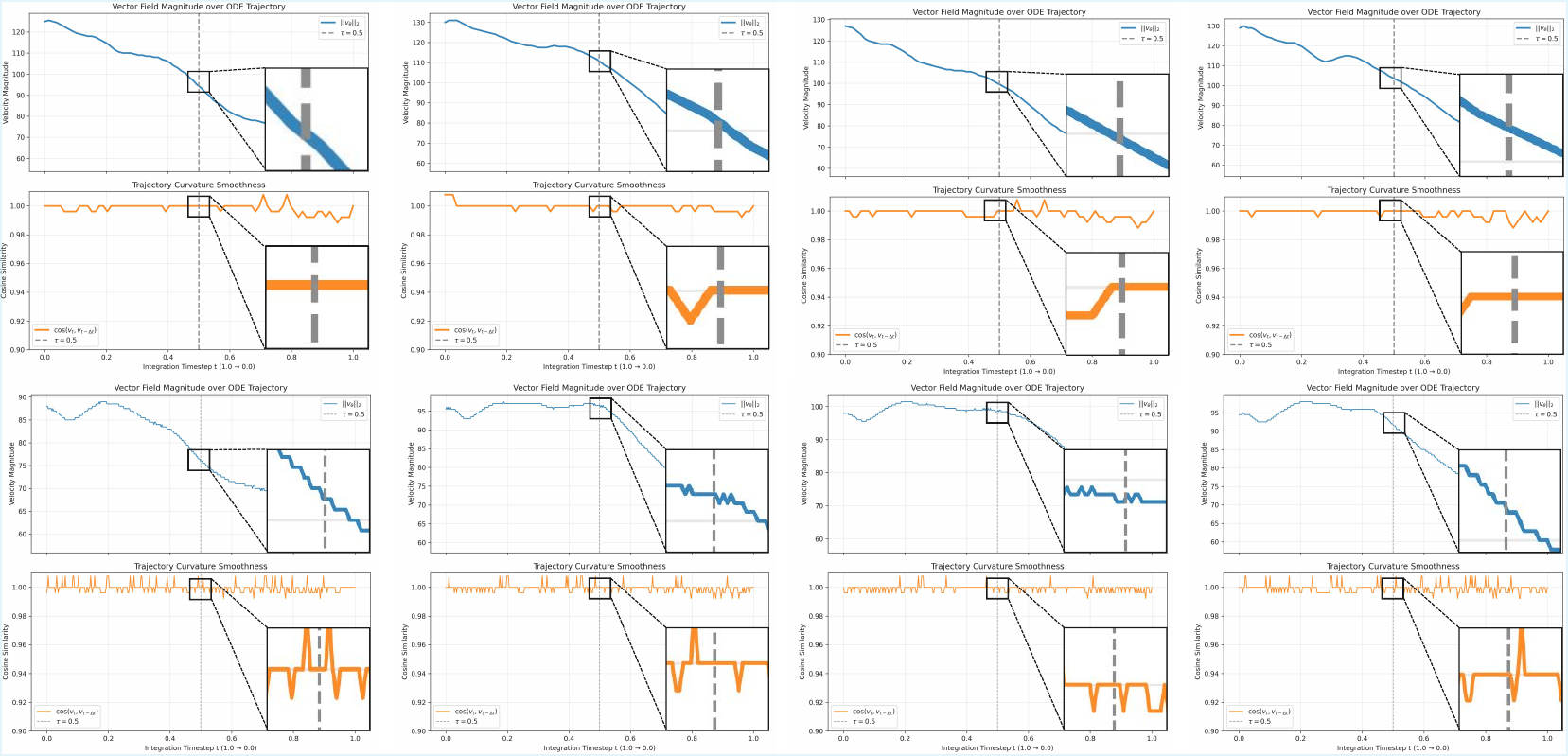}
\vspace{-20pt}
\caption{ODE Stability Validation. The top 2 rows: early-stage trajectories monitored every 200 training steps with a coarse integration resolution of 80 steps. The bottom 2 rows: late-stage trajectories evaluated every 2000 training steps using a fine-grained integration resolution of 400 steps. In both settings, the vector field magnitude (blue) and trajectory curvature (orange) remain highly stable across the truncation boundary $\tau=0.5$ (dashed line), validating the absence of abrupt discontinuities.}
\label{fig:ODE}
\vspace{-10pt}
\end{figure*}

\section{Dive into Timestep-Truncation Mechanism}
\label{sec:timestep_truncation}

In the main manuscript, we introduced a Timestep-Truncation Mechanism ($\tau=0.5$) within the semantic loss $\mathcal{L}_{sem}$ to decouple semantic alignment from physical manifold reconstruction. While this mechanism effectively decouples the training objectives, it imposes an explicit hard boundary at $t=\tau$ where the loss function undergoes an abrupt transition. This structural shift inherently risks inducing discontinuities in the learned vector field $v_\theta(x_t, T, t)$. Such potential discontinuities could drastically amplify discretization errors and jeopardize the numerical stability of the Ordinary Differential Equation (ODE) solvers. Theoretically, the continuous generative process is governed by the probability flow ODE:
\begin{equation}
    \mathrm{d}x_t = v_\theta(x_t, T, t) \mathrm{d}t, \quad t \in [0, 1]
\end{equation}
According to the Picard-Lindel\"{o}f theorem \cite{siegmund2016generalized}, ensuring the existence and stability of the ODE trajectory necessitates that the vector field $v_\theta$ be Lipschitz continuous \cite{goldstein1977optimization} with respect to $x_t$ and continuous with respect to $t$. Although our training loss incorporates a hard truncation $\mathbb{I}(t > \tau)$, the neural network $v_\theta$ inherently acts as a continuous prior. The continuous timestep embeddings and the smooth activation functions implicitly prevent step-changes in the output manifold. Furthermore, the global expectation optimization over shared parameters enables the network to internalize the dual piecewise constraints into a unified, smooth global flow. To validate this continuity, we conducted a rigorous stability analysis. Specifically, we periodically monitor the inference trajectory during training by performing explicit Euler integration over $t \in [0, 1]$. At each timestep, we record the numerical behavior of the vector field to ensure its integration reliability, focusing on two critical stability metrics: \noindent\textbf{Vector Field Magnitude ($||v_\theta||_2$)} measures the "velocity" of the transformation. A smooth magnitude curve indicates the absence of sudden acceleration or gradient explosions in the latent space. \noindent\textbf{Trajectory Curvature ($\cos(v_t, v_{t-\Delta t})$)} computes the cosine similarity between consecutive velocity vectors. A value approaching 1.0 signifies a highly linear and stable integration path without abrupt directional shifts. As illustrated in Fig. \ref{fig:ODE}, both metrics exhibit exceptionally stable behavior around the threshold $\tau=0.5$. The magnitude curve shows no violent spikes, and the cosine similarity remains consistently high throughout the entire process. This empirical evidence confirms that the truncation mechanism solely alters the weight allocation during training, without introducing vector field discontinuities. Consequently, it guarantees that ODE solvers can integrate the path precisely without discretization error explosion, thereby preserving the stringent structural integrity of the fused medical images.